\documentclass{article}

\usepackage{PRIMEarxiv}

\usepackage[utf8]{inputenc} 
\usepackage[T1]{fontenc}    
\usepackage{hyperref}       
\usepackage{url}            
\usepackage{booktabs}       
\usepackage{amsfonts}       
\usepackage{nicefrac}       
\usepackage{microtype}      
\usepackage{lipsum}
\usepackage{fancyhdr}       
\usepackage{graphicx}       
\usepackage[linesnumbered,ruled,vlined]{algorithm2e}
\graphicspath{{media/}}     
\usepackage{subcaption}
\usepackage{amsmath}
\newtheorem{theorem}{Theorem}
\newtheorem{definition}{Definition}
\newtheorem{assumption}{Assumption}
\newtheorem{lemma}{Lemma}

\title{Sepsyn-OLCP: An Online Learning-based Framework for Early Sepsis Prediction with Uncertainty Quantification using Conformal Prediction
}

\author{
  Anni Zhou, Raheem Beyah \\
  School of Electrical and Computer Engineering\\
  Georgia Institute of Technology \\
  \texttt{\{azhou60, ab207\}@gatech.edu} \\
   \And
  Rishikesan Kamaleswaran \\
  School of Medicine, Department of Surgery\\
  Duke University \\
  \texttt{r.kamaleswaran@duke.edu} \\
  \AND
  Yao Xie \\
  H. Milton Stewart School of Industrial and Systems Engineering\\
  Georgia Institute of Technology \\
  \texttt{yxie77@gatech.edu} \\
}

\begin{document}
\maketitle

\begin{abstract}
Sepsis is a life-threatening syndrome with high morbidity and mortality in hospitals. Early prediction of sepsis plays a crucial role in facilitating early interventions for septic patients. However, early sepsis prediction systems with uncertainty quantification and adaptive learning are scarce. 
This paper proposes Sepsyn-OLCP, a novel online learning algorithm for early sepsis prediction by integrating conformal prediction for uncertainty quantification and Bayesian bandits for adaptive decision-making. By combining the robustness of Bayesian models with the statistical uncertainty guarantees of conformal prediction methodologies, this algorithm delivers   accurate and trustworthy predictions, addressing the critical need for reliable and adaptive systems in high-stakes healthcare applications such as early sepsis prediction.
We evaluate the performance of Sepsyn-OLCP in terms of regret in stochastic bandit setting, the area under the receiver operating characteristic curve (AUROC), and F-measure. Our results show that Sepsyn-OLCP outperforms existing individual models, increasing AUROC of a neural network from 0.64 to 0.73 without retraining and high computational costs. And the model selection policy converges to the optimal strategy in the long run.
We propose a novel reinforcement learning-based framework integrated with conformal prediction techniques to provide uncertainty quantification for early sepsis prediction. The proposed methodology delivers accurate and trustworthy predictions, addressing a critical need in high-stakes healthcare applications like early sepsis prediction.
\end{abstract}


 \section{Introduction}

Sepsis is a severe and often life-threatening condition defined by organ dysfunction arising from an uncontrolled host response to infection \cite{sepsis3, physionetChallenge}. It can lead to significant organ damage, tissue injury, or even death if not promptly addressed. While bacterial infections are the most common cause, viral infections, including those caused by Covid19, can also lead to sepsis. Among critically ill patients in intensive care units (ICUs), sepsis remains a major contributor to both morbidity and mortality worldwide \cite{coopersmith2018surviving, weiss2020surviving}.

The burden of sepsis is not just significant, it is staggering on a global scale. A recent study estimated that sepsis contributed to nearly 20\% of all deaths worldwide in 2017, with the high incidence and mortality rates observed in low-resource settings. Vulnerable populations—including young children, the elderly, and individuals with compromised immune systems—bear a disproportionate share of this burden. In high-resource settings like the United States, sepsis similarly represents a critical health concern. At least 1.7 million adults in the U.S. develop sepsis annually, and one in three hospital deaths is related to sepsis according to the Centers for Disease Control and Prevention (CDC). Furthermore, the financial costs associated with sepsis are rising. From 2012 to 2018, the number of hospital admissions for sepsis-related conditions among fee-for-service beneficiaries increased from approximately 812,000 to over 1.1 million, underscoring its growing prevalence and economic impact.

Despite advancements in critical care, sepsis outcomes remain concerning. Survival rates for severe sepsis are approximately 70\%, but nearly half of survivors face long-term complications, collectively referred to as post-sepsis syndrome. These complications often include physical disabilities, cognitive impairments, and psychological challenges, underscoring the lasting impact of the condition. Without a definitive cure, early detection and timely intervention remain the most effective strategies for improving survival rates and minimizing long-term disabilities\footnote{\url{https://www.sepsis.org/sepsis-basics/what-is-sepsis/}}.

Sepsis progresses rapidly, with a narrow window for effective treatment. Delayed diagnosis can lead to irreversible organ failure, increased mortality risk, and extended hospital stays, all of which amplify the strain on healthcare systems. Current clinical practices rely heavily on detecting sepsis after its symptoms have become apparent, which limits the ability to intervene during its early stages. This delay not only worsens patient outcomes but also imposes substantial economic costs on public health systems. Therefore, early identification of sepsis remains critical for effective patient management, timely intervention, and improved survival outcomes. 


Recently, machine learning (ML) has emerged as a promising alternative for predicting sepsis onset by effectively leveraging clinical data from electronic health records (EHRs). ML methods have shown substantial improvements over conventional scoring systems by incorporating diverse patient data, including demographics, vital signs, and laboratory results \cite{barton2019evaluation, delahanty2019development, nemati2018interpretable}. Barton et al. (2019) demonstrated that ML models significantly outperform traditional methods (e.g., SOFA, qSOFA) in detecting sepsis up to 48 hours before clinical onset, achieving AUROC scores up to 0.88 \cite{barton2019evaluation}. Similarly, Delahanty et al. (2019) developed the Risk of Sepsis (RoS) scoring system, utilizing gradient boosting, reporting excellent performance with AUROC values between 0.93 and 0.97 \cite{delahanty2019development}. Nemati et al. (2018) and Mao et al. (2018) further illustrated the efficacy and interpretability of Random Forest and XGBoost algorithms, achieving strong predictive accuracy in clinical environments \cite{nemati2018interpretable, mao2018multicentre}.

Deep learning models, notably recurrent neural networks (RNNs) and convolutional neural networks (CNNs), have further enhanced sepsis prediction by capturing complex temporal dependencies in EHR data \cite{scherpf2019predicting, kok2020automated}. Kok et al. (2020) demonstrated superior performance with temporal convolutional networks, achieving high predictive accuracy and robustness \cite{kok2020automated}. Rafiei et al. (2021) introduced an advanced CNN-LSTM hybrid model, accurately predicting sepsis onset up to 12 hours in advance with remarkable AUROC values \cite{rafiei2021ssp}. Such approaches underscore the potential of ML and DL methods to revolutionize clinical sepsis detection practices.

Despite significant advances, existing models frequently lack mechanisms for uncertainty quantification and adaptive learning, critical for dynamic clinical environments where data distributions evolve \cite{zhou2021onai}. To address these challenges, we propose\textit{ \textbf{Sepsyn-OLCP}}, an online learning-based framework integrating Bayesian bandit approaches with conformal prediction methods, robustly quantifying prediction uncertainty while dynamically adapting to incoming data. By combining Bayesian modeling, reinforcement learning and conformal prediction techniques, our framework continuously updates its decision-making strategies based on real-time patient data.
The contribution of this paper can be highlighted as:

\begin{itemize}
    \item We proposed Sepsyn-OLCP, a novel methodology that rigorously evaluates the predictive performance and clinical utility for early sepsis prediction, comparing it directly against established ML benchmarks. 
    \item Our comprehensive experiments demonstrated that our approach has superior predictive accuracy, robust uncertainty quantification, and enhanced adaptability, ultimately facilitating improved patient outcomes in sepsis management.
     \item Our proposed framework can improve prediction accuracy and uncertainty quantification at low cost without retraining the model by incorporating a novel ensemble-based conformal prediction methodology into a Bayesian modeling-based reinforcement learning model. 
     \item Our methodology can provide personalized recommendations of AI clinicians based on patients' contextual information. 
\end{itemize}

 
The remainder of the paper is organized as follows: Section \ref{sec:mat} demonstrates the details of the problem formulation, and mathematical foundation of the proposed framework. Section \ref{sec:res} provides the details of dataset description, data preprocessing methods, experimental setup and experimental results, and Section \ref{sec:disc} discusses and analyzes the results. Finally, we conclude the study in Section \ref{sec:conc}.

\section{Materials and methods}
\label{sec:mat}
We model the decision process in the early sepsis prediction problem as an AI expert competing system, where the expert with the highest utility score/reward will win the game and be recommended to the current patient. There are several main components in the problem:
\begin{itemize}
    \item \textbf{Patient.} A group of patients $\{P_1,P_2, \cdots,P_t,\cdots, P_T\}$ arrive sequentially at the hospital. $t \in \{1,2,\cdots\,T\}$ and $t$ is associated with a unique timestamp. For simplicity, the patient arriving at each round is considered unique, even if the patient is the same person.
    \item \textbf{AI Expert.} A group of AI experts $\mathcal{E}_K = \{A_1,A_2,\cdots,A_K\}$ provide ML as a Service (MLaS) aiming to help clinicians/hospitals analyze EMRs. In this system, we assume that there are $K$ AI experts. Each AI expert is assumed to be pre-trained using the same training dataset.
    \item \textbf{Trusted third party (TTP).} A TTP can be an entity (e.g., a hospital) that has hired $K$ experts to solve the early sepsis prediction task. 
\end{itemize}
 
\begin{figure}[htb!]
    \centering
    \includegraphics[width=\linewidth]{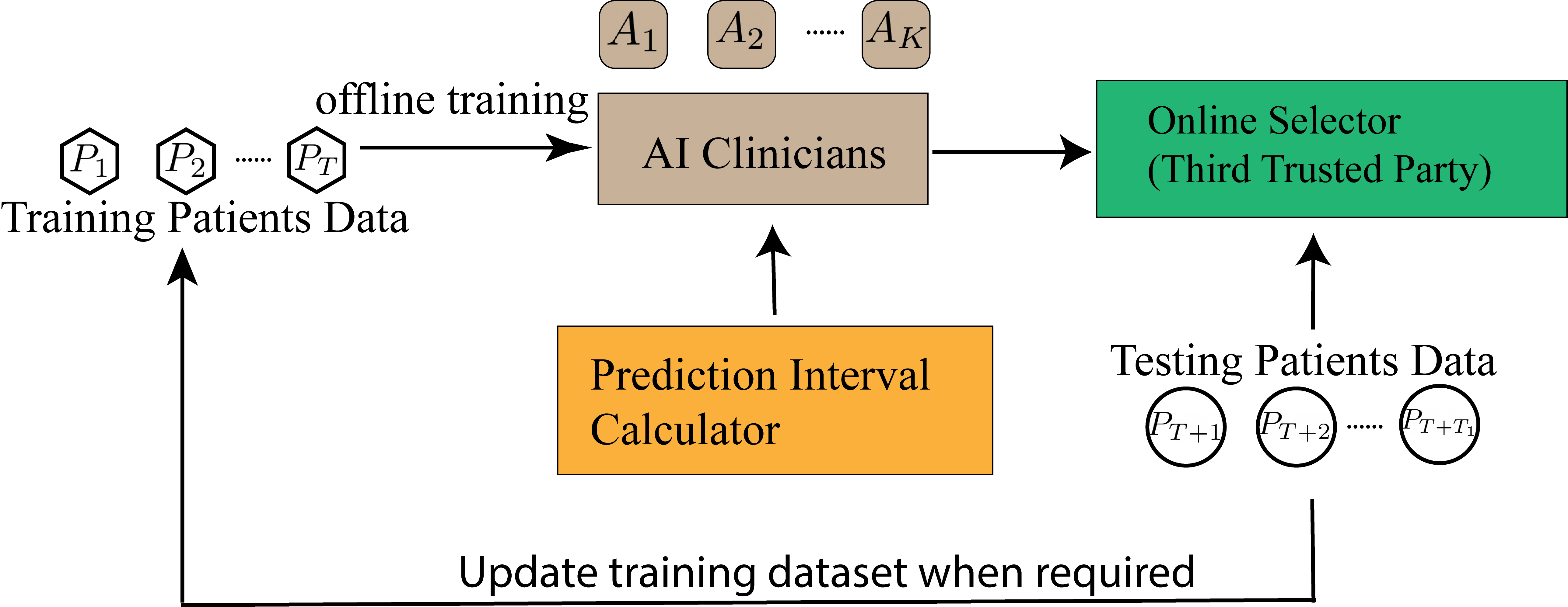}
    \caption{System Overview of Sepsyn-OLCP.}
    \label{fig:cpbandit_highlevel}
\end{figure}
In this paper, all the objects in the framework are assumed to be benign, and a certain protocol is observed to protect sensitive information from patients. However, components in the system might be attacked by external malicious attackers. So, the TTP should provide all EMRs during the online learning process and offline training.

As shown in Fig. \ref{fig:cpbandit_highlevel}), our framework has two main modules: an  \textit{Online Selector} which selects the best promising AI expert for the current patient based on the patient's contextual information (e.g., demographics, lab values, vital signs, etc.), and a \textit{Prediction Interval Calculator} which provides uncertainty quantification for predictions of each AI expert. 

Then we will explain the details of theoretical foundations and core methodologies utilized in the and \textit{Prediction Interval Calculator} in our framework (see Fig. \ref{fig:cpbandit_highlevel}).
The Online Selector is designed based on \textit{BayesGap}, which is a gap-based solution for the Bayesian optimization problem under a bandit setting \cite{hoffman2014correlation}. We focus on the \textit{fixed budget setting} \cite{gabillon2012best} in predicting the sepsis onset time considering the nature of the big healthcare dataset and the complexity of early sepsis prediction problems. The Prediction Interval Calculator adapts from \textbf{\textit{EnbPI}} \cite{EnbPI}. In the end, we consider the two perspectives together and integrate them into a new framework. We also provide new theoretical analysis based on the similar assumptions in \cite{hoffman2014correlation} and \cite{EnbPI}.



\subsection{Online Selector}

Specifically, we model the selection of the best AI expert as a contextual multi-armed bandit problem with $K$ arms over time horizon $T$ and context set $x_{t,a_t}\in\mathcal{X}$, where $k=1,\cdots, K$. We denote $\mathcal{A} =\{1,2,\cdots K \}$ as the arm set.
At each time $t$ we first observe the context for each arm, then choose an arm $a_t\in\mathcal{A} $ and receive the reward. 
And the reward of pulling $a_t \in \mathcal{A}$ at round $t$ is denoted as:
\begin{equation}\label{eq:reward_generation}
    r_{t,a_t} = f_{a_t}(x_{t,a_t}) + \epsilon_{t,a_t}, 
\end{equation}
where $f_1,\cdots,f_K: \mathcal{X} \to \mathbb{R}$ are unknown and $\epsilon_{t,a_t}$ is the noise drawn from an unknown distribution $F_{a_t}$. We do not require the noises $\epsilon_{t,k},t\in[T],k\in[K]$ to be independent. The context $x_{t,k}$ can be either exogenous data or the history of reward on arm $k$. In this paper, $x_{t,k}$ is the electrical health records (e.g., lab values, vital signs, demographics, .etc) of the ICU patients recorded on an hourly basis. 

The average regret up till round $T$ is defined as the average difference of expected reward between the optimal arms the selected arms at each round $t$:
\begin{equation}\label{eq:cul_regret}
    R_T = \frac{1}{T}\sum_{t=1}^T \left(f_{a_t^*}(x_{t,a_t^*})  + \mathbb{E}[\epsilon_{t,a_t^*}]\right)- \frac{1}{T} \sum_{t=1}^T \left(f_{a_t}(x_{t,a_t})+ \mathbb{E}[\epsilon_{t,a_t}]\right),
\end{equation}
where $a_t^* = \arg\max_{k\in[K]}f_{k}(x_{t,k})  + \mathbb{E} [\epsilon_{t,k}]$, $t=1,\cdots,T$.

We assume the reward distribution for each arm depends on unknown parameters $\psi \in \Psi$ that are shared among all arms. We denote the reward distribution for arm $j$ as $\kappa_j(\cdot|\psi)$. In a Bayesian framework for the bandit problem, we assume that the parameters $\psi$ follow a prior density $\pi_0(\cdot)$. The posterior density of the parameters $\psi$ after $s - 1$ rounds can be formalized as

\begin{equation}
\pi_s(\psi) \propto \pi_0(\psi) \prod_{m < s} \kappa_{a_m}(y_m|\psi).
\label{eq:posterior_density_alt}
\end{equation}
Eq. \ref{eq:posterior_density_alt} shows that by choosing arm $a_m$ at each time step $m$, we gather information about $\psi$ only indirectly through the likelihood of these parameters, given the observed rewards $y_m$. This also extends to the scenario where the arms are uncorrelated. If the rewards for each arm $j$ only depend on a set of parameters (or a specific parameter) $\psi_j$, then at time $s$, the posterior for $\psi_j$ would be influenced only by the rounds when arm $j$ was previously selected.

However, the posterior distribution of the parameters $\psi$ is not our main interest. Instead, our focus is on the expected reward for each arm under these parameters, which we define as
\begin{equation}
\nu_j = \mathbb{E}[Y|\psi] = \int_y \kappa_j(y|\psi) \,dy.
\label{eq:expected_reward_alt}
\end{equation}
Although the exact value of $\psi$ is unknown, we have access to its posterior distribution, $\pi_s(\psi)$. This distribution induces a marginal distribution over $\nu_j$, represented as $\rho_{j,s}(\nu_j)$. 
In a gap-based bandit problem \cite{gabillon2012best,hoffman2014correlation}, we aim to establish upper and lower confidence bounds that hold with high probability, allowing us to design acquisition functions that can strike a balance between exploration and exploitation. 

Assuming that each arm $j$ is associated with a feature vector $c_j \in \mathbb{R}^d$, then we can formulate the rewards for selecting arm $j$ to follow a normal distribution as follows

\begin{equation}
\kappa (y|\psi) \sim \mathcal{N}(y; c_j^T \psi, \sigma^2),
\label{eq:normal_distribution_alt}
\end{equation}
in which $\sigma^2$ is the variance and $\psi \in \mathbb{R}^d$ the unknown parameter. 
The rewards $\mathbf{v}_j$ for each arm $j$ are conditionally independent given $\psi$, but share a marginal dependency through $\psi$ when $\psi$ is unknown.
The specific dependence of the rewards $\mathbf{v}$ is defined by the structure of the vectors $\mathbf{v}_j$. If we place a prior on $\psi \sim \mathcal{N}(0, \tau^2 I)$, we can determine a posterior distribution over the unknown parameter $\psi$.
Specifically, let the matrix $G \in \mathbb{R}^{K \times K}$ represent the covariance of a Gaussian process (GP) prior. 
The matrix $\mathbf{C} = [\mathbf{c}_1 \dots \mathbf{c}_K]^T$ can be constructed as follows:
\begin{equation}
\mathbf{C} = V D^{1/2}, \quad \text{where } G = V D V^T,
\label{eq:design_matrix_alt}
\end{equation}
where each row of $\mathbf{C}$ is a vector $\mathbf{c}_j,$ where $j\in\{1,2,\cdots, K\}$. The matrix $\mathbf{C}$ is essential in setting up the observation model (see \ref{eq:normal_distribution_alt}). 
In practical scenarios, popular Bayesian optimization methods often consider either finite action grids or apply discretization to the space of possible actions. In this paper, we focus on the finite and discrete action spaces.

Let $\mathbf{Y}_t = \{y_1, y_2, \cdots, y_t\}_{t=1}^T$ represent the observed rewards up to round $t$, and let $\mathbf{C}_t = \{\mathbf{c}_{a_1},\mathbf{c}_{a_2}, \cdots,\mathbf{c}_{a_t}\}_{t=1}^T$ be the sequence of feature vectors correlated with the selected arms, where $a_t$ is the arm selected at round $t$.
Then, we can write the posterior at round $t$ as follows:
\begin{equation}
\pi_t(\psi) = \mathcal{N}(\psi;\hat{\psi}_t, \hat{\Sigma}_t), 
\label{eq:pi_t}
\end{equation}
where
\begin{equation}
\Sigma_t^{-1} = \sigma^{-2} \mathbf{C}_t^T \mathbf{C}_t + \tau^{-2} I,
\label{eq:posterior_covariance_alt}
\end{equation}
and
\begin{equation}
\hat{\psi}_t = \sigma^{-2} \Sigma_t \mathbf{C}_t^T \mathbf{Y}_t.
\label{eq:posterior_mean_alt}
\end{equation}
Based on the above formulation, the expected reward correlated with arm $k$ can be derived as a marginally normal distribution:
\begin{equation}
\rho_{k,t}(\nu_k) = \mathcal{N}(\nu_k;\hat{\nu}_{k,t}, \hat{\sigma}_{k,t}), 
\label{eq:=rho_kt}
\end{equation}
in which the mean $\hat{\nu}_{k,t} = c_j^T \hat{\psi}_t$ and the variance $ \hat{\sigma}_{k,t}^2=c_j^T \hat{\Sigma}_t c_j$. In addition, the predictive distribution of rewards associated with arm $k$ is normal with mean $\hat{\nu}_{k,t}$ and variance $\hat{\sigma}_{k,t}^2 + \ \sigma^2$.

At the start of round $t$, we assume the decision-maker has high-probability bounds, denoted as $U_k(t)$ (upper bounds) and $L_k(t)$ (lower bounds), with the unknown mean $\mu_k$ of each arm.
For the simplicity of theoretical analysis as in \cite{hoffman2014correlation}, we define the upper bounds and lower bounds in terms of the mean and standard deviation, specifically $\hat{\mu}_{k,t} \pm \beta \hat{\sigma}_{k,t}$. These bounds create a confidence interval diameter $s_k(t) = U_k(t) - L_k(t) = 2 \beta \hat{\sigma}_{k,t}$. 
Note that this approach can accommodate more general bounds. In this work, we focus on the Gaussian arm settings.

With these bounds on the mean reward for each arm, we can now define the gap of arm $k$ at round $t$ as follows:
\begin{equation} \label{eq:gap_quantity}
B_k(t) = \max_{i \neq k} U_i(t) - L_k(t),
\end{equation}
which represents the gap between the lower bound of arm $k$ with the highest upper bound among all other arms. 
\ref{eq:gap_quantity} ultimately provides an upper bound on the simple regret and serves to guide the exploration strategy. Then, we will focus on two essential arms in gap-based bandit problems:
\begin{equation} \label{eq:min_gap_arm}
J(t) = \arg \min_{k \in A} B_k(t)
\end{equation}
and
\begin{equation} \label{eq:max_upper_bound}
j(t) = \arg \max_{k \neq J(t)} U_k(t).
\end{equation}
And, the exploration strategy is defined as choosing from $k \in \{j(t), J(t)\}$ that maximizes the confidence diameter:
\begin{equation} \label{eq:exploration_strategy}
a_t = \arg \max_{k \in \{j(t), J(t)\}} s_k(t).
\end{equation}
The intuition of this strategy is that we will select either the arm that minimizes the bound on simple regret (i.e., $J(t)$) or the best "runner-up" arm (i.e., $j(t)$). The arm with the highest uncertainty, which is expected to provide the most information, will be chosen from $\{j(t), J(t)\}$.
Then, the final arm selection strategy is defined as: 
\begin{equation} \label{eq:recommendation_strategy}
\Omega_T = J \left( \arg \min_{t \leq T} B_{J(t)}(t) \right).
\end{equation}

\subsection{Prediction Interval Calculator}
Suppose that the data \((x_i, y_i)\) observed is generated from the following model:
\begin{equation}\label{eq:Yt}
    Y_i = f(X_i) + \xi_i, \quad i = 1, 2, \dots,
\end{equation}
in which \(f : \mathbb{R}^d \to \mathbb{R}\) is an unknown model; \(d\) is the dimension of the feature vector; and \(\xi_i\) is generated from a continuous cumulative distribution function (CDF) \(F_i\).
The first \(n\) samples \(\{(x_i, y_i)\}_{is=1}^n\), i.e., training data or initial state of the random process, are assumed to be observable.  

The ultimate goal of the conformal prediction algorithm is to construct a sequence of prediction intervals with a certain coverage guarantee and to make the interval width as narrow as possible. 
First, we obtained a well-trained model \(\hat{f}\) using \(n\) training samples.
Then we construct prediction intervals \(\{\hat{C}_{j}^{\alpha}\}_{j=n+1}^{n+bs}\) for \(\{Y_{j}\}_{j=n+1}^{n+bs}\), in which \(bs \geq 1\). \(\alpha\) is the significance level. The batch size \(bs\) defines how many steps we want to look ahead.

After new samples \(\{(x_{j}, y_{j})\}_{j=n+1}^{n+bs}\) become available, the pre-trained \(\hat{f}\) is deployed on new samples and the most recent \(n\) samples are used to produce prediction intervals for \(\{Y_j\}_{j = n + bs + 1}\) onward without re-training the model on new data.

As in a standard conformal prediction problem, we consider two types of coverage guarantees. 
\begin{definition}
\label{def:conditional}
\textbf{[Conditional Coverage Guarantee]}
The conditional coverage guarantee ensures that each prediction interval \(\hat{C}_j^{\alpha}, \forall j > n\) satisfies:
\begin{equation}\label{eq:coverage1}
    P(Y_j \in \hat{C}_j^{\alpha} | X_j = x_j) \geq 1 - \alpha.  
\end{equation}
\end{definition}

\begin{definition}\label{def:marginal}
\textbf{[Marginal Coverage Guarantee]}
The marginal coverage guarantee ensures that each prediction interval \(\hat{C}_j^{\alpha}, \forall j > n\) satisfies:
\begin{equation}\label{eq:coverage2}
    P(Y_j \in \hat{C}_j^{\alpha}) \geq 1 - \alpha.  
\end{equation}
\end{definition}

A prediction interval is called conditionally or marginally valid if it can achieve \ref{eq:coverage1} or \ref{eq:coverage2}, respectively. \ref{eq:coverage2} is satisfied whenever the data is exchangeable using split conformal prediction \cite{papadopoulos2007conformal}.
In our early sepsis prediction scenario, we assume that a clinician
receives a prediction interval for the probability that the current patient develops sepsis on an hourly basis. If the interval satisfying \ref{eq:coverage2}
is evenly distributed in all patients in different age groups, it may not exactly satisfy \ref{eq:coverage1} for the current
patient.
In fact, it is impossible to satisfy \ref{eq:coverage1} even for exchangeable data without
additional assumptions \cite{foygel2021limits, EnbPI}. Generally, it is challenging to ensure any of the aforementioned coverage guarantees under complex data dependency without specific distributional assumptions \cite{EnbPI}.
Taking into account these challenges, we can bound the worst-case gap in \ref{eq:coverage1} and \ref{eq:coverage2} under certain assumptions by adapting the conformal prediction algorithm introduced in \cite{EnbPI}.
\subsubsection{Intuition for Construction of Prediction Intervals} 

We construct our prediction intervals following the similar steps demonstrated in \cite{EnbPI,sesia2021conformal}. The oracle prediction interval \(C_j^{\alpha}\) contains \(Y_j\) with an exact conditional coverage at \(1 - \alpha\) and is the narrowest among all possible conditionally valid prediction intervals.
In an oracle prediction scenario, we assume perfect knowledge of \(f\) and \(F_i\) in \ref{eq:Yt}. 
We denote \(F_{i,Y}\) as the CDF of \(Y_i\) conditioning on \(X_i = x_i\), then we have
\begin{equation} \label{eq:F_{i,Y}(y)}
F_{i,Y}(y) = \mathbb{P}(Y_i \leq y | X_i = i_t)= \mathbb{P}(\xi_i \leq y - f(x_i)) = F_i(y - f(x_i)).
\end{equation}
Based on \ref{eq:coverage1}, our goal is to create an interval such that the probability of $Y_i$ falls within this interval is exactly $1-\alpha$. 
First, we need to find the quantile function as a general conformal prediction algorithm does. We denote $F_{i,Y}^{-1}(\beta)$ as the inverse cumulative distribution function of $Y_i$. $F_{i,Y}^{-1}(\beta)$ finds the value of $y$ such that the probability $\mathbb{P}(Y_i \leq y | X_i = x_i)$ is equal to $\beta$. The inverse CDF tells at which value of $Y_i$ the cumulative probability equals $\beta$. This helps us define the boundaries of our prediction interval.

Then, we aim to construct ensemble intervals.
Interval $[F_{i,Y}^{-1}(\beta), F_{i,Y}^{-1}(1-\alpha+\beta)]$ is designed to capture a specific portion of the total probability. The upper bound of the interval is $F_{i,Y}^{-1}(1-\alpha+\beta)$, where $\beta \in [1, \alpha]$. This upper bound can ensure that the total probability contained within the interval is $1-\alpha$.
   
\subsubsection{Theoretical Analysis of the Total Probability}
By definition, $F_{i,Y}$ is the CDF of $Y_i$, i.e., it describes the probability that $Y_i$ takes on value less than or equal to a give number.
To construct a prediction interval with desired probability coverage, we choose $1-\alpha+\beta$ and $\beta$ so that the difference in cumulative probabilities between is $1-\alpha$.
Mathematically, the probability that $Y_i$ falls within $[F_{i,Y}^{-1}(\beta), F_{i,Y}^{-1}(1-\alpha+\beta)]$ is:
\begin{equation}\label{eq:proof_F_interval}
\begin{split}
& \mathbb{P}(Y_i \in [F_{i,Y}^{-1}(\beta), F_{i,Y}^{-1}(1-\alpha+\beta)]| X_i = x_i)  \\
    &= F_{i,Y}(F_{i,Y}^{-1}(1-\alpha+\beta))-F_{i,Y}(F_{i,Y}^{-1}(\beta)) \\
    &=1-\alpha.
\end{split}
\end{equation}
 
In conclusion, now for any \(\beta \in [0, \alpha]\), we have
\[
\mathbb{P}(Y_i \in [F_{i,Y}^{-1}(\beta), F_{i,Y}^{-1}(1 - \alpha + \beta)] | X_i = x_i) = 1 - \alpha,
\]
where \(F_{i,Y}^{-1}(\beta) := \inf\{y : F_{i,Y}(y) \geq \beta\}\). 
Let \(y_{\beta} = F_{i,Y}^{-1}(\beta)\), then we have
\[
y_{\beta} = f(x_i) + F_i^{-1}(\beta),
\]
which allows us to find \(C_i^{\alpha}\) – the oracle prediction interval with the narrowest width:
\begin{equation}\label{eq:hat_c_alpha}
C_i^{\alpha} = [f(x_i) + F_i^{-1}(\beta^*), f(x_i) + F_i^{-1}(1 - \alpha + \beta^*)],
\end{equation}
where
\[
\beta^* = \arg\min_{\beta \in[0, \alpha]}(F_{i}^{-1}(1 - \alpha + \beta) -F_{i}^{-1}(\beta)).
\]
After constructing the oracle interval $C_i^{\alpha}$, next we need to figure out a way to approximate $C_i^{\alpha}$ well as in a standard conformal prediction algorithm \cite{angelopoulos2021gentle}.

\subsubsection{Approximation of Prediction Intervals}

In a conformal prediction problem, we need to split the training data into two parts: 1) the first partition is used to estimate $f$; 2) the second partition is used to obtain prediction residuals which is required to calculate the final conformal prediction interval. There is a trade-off between using as much data as possible to approximate $f$ and using the quantile of prediction residuals to approximate the prediction interval indicated in \ref{eq:hat_c_alpha}. 
On the one hand, we want to use as much data as possible to train the estimator $\hat{f}$. On the other hand, we want the quantile of prediction residuals to be well
approximate the tails of $F_i^{-1}(\beta^*)$ and $F_i^{-1}(1 - \alpha + \beta^*)$.

If we use all training data to approximate $f$, we might have an over-fitting problem. If we train only on a subset of training data to avoid over-fitting and calculate the prediction residuals on the rest \cite{papadopoulos2007conformal}, we will get worse approximation of $F_i^{-1}(\beta^*)$ and $F_i^{-1}(1 - \alpha + \beta^*)$. So there is a dilemma. 

To solve this dilemma, we will use the well-known Leave One Out (LOO) estimator, where the $i$-th residua is obtained by training the $i$-th LOO estimator on all except the $i$-th training entry $(x_i,y_i)$ so that the LOO estimator is not overfitted on that datum. Then, repeating over $T$ training data yields $T$ LOO estimators with good predictive power and $T$ residuals to calibrate the prediction intervals well.

The LOO methodology can strike a good balance between the approximating $f$ and well calculating prediction residuals \cite{barber2021predictive}.
However, the LOO methodology is known to be extremely computationally expensive since we need to retrain the model. To avoid the high computational complexity, we use the computationally efficient method in \cite{EnbPI} which utilizes the pre-trained models to obtain the LOO estimators.



\subsubsection{The Final Ensemble Prediction Interval}
We assume that the first $n$ data points $\{(x_i, y_i)\}_{i=1}^n$ are observable.
Following the aforementioned intuitions in the previous subsection, we approximate our conformal prediction interval as follows
\begin{equation}
\begin{split}
      \hat{C}_i^\alpha =
     &[\hat{f}_{-i}(x_i) + \beta \text{ quantile of } \{\hat{\xi}_j\}_{j=i-1}^{i-n}, \\
      &\quad \hat{f}_{-i}(x_i) + (1 - \alpha + \beta) \text{ quantile of } \{\hat{\xi}_j\}_{j=i-1}^{i-n}],  
\end{split}
\end{equation}
in which $\hat{f}_{-i}$ is defined as the $i$-th ``leave-one-out'' estimator of $f$. In other words, $f_{-i}$ is not trained on the $i$-th entry $(x_i, y_i)$ and may include the remaining $n-1$ training data points. 

The LOO prediction residual $\hat{\xi}_i$ is calculated as:
\begin{equation}\label{eq:hat_xi}
    \hat{\xi}_i := y_i - \hat{f}_{-i}(x_i),
\end{equation}
and the corresponding $\hat{\beta}$ are calculated as follows:
\begin{equation}
\beta^* = \arg\min_{\beta \in [0, \alpha]} \left( (1 - \alpha + \beta) \text{ quantile of } \{\hat{\xi}_j\}_{j=i-1}^{i-n} - 
        \beta \text{ quantile of } \{\hat{\xi}_j\}_{j=i-1}^{i-n} \right).
\label{eq:hat_beta}
\end{equation}


\subsection{System Workflows of Sepsyn-OLCP}

\begin{figure*}[h]
    \centering
    \includegraphics[width=\linewidth]{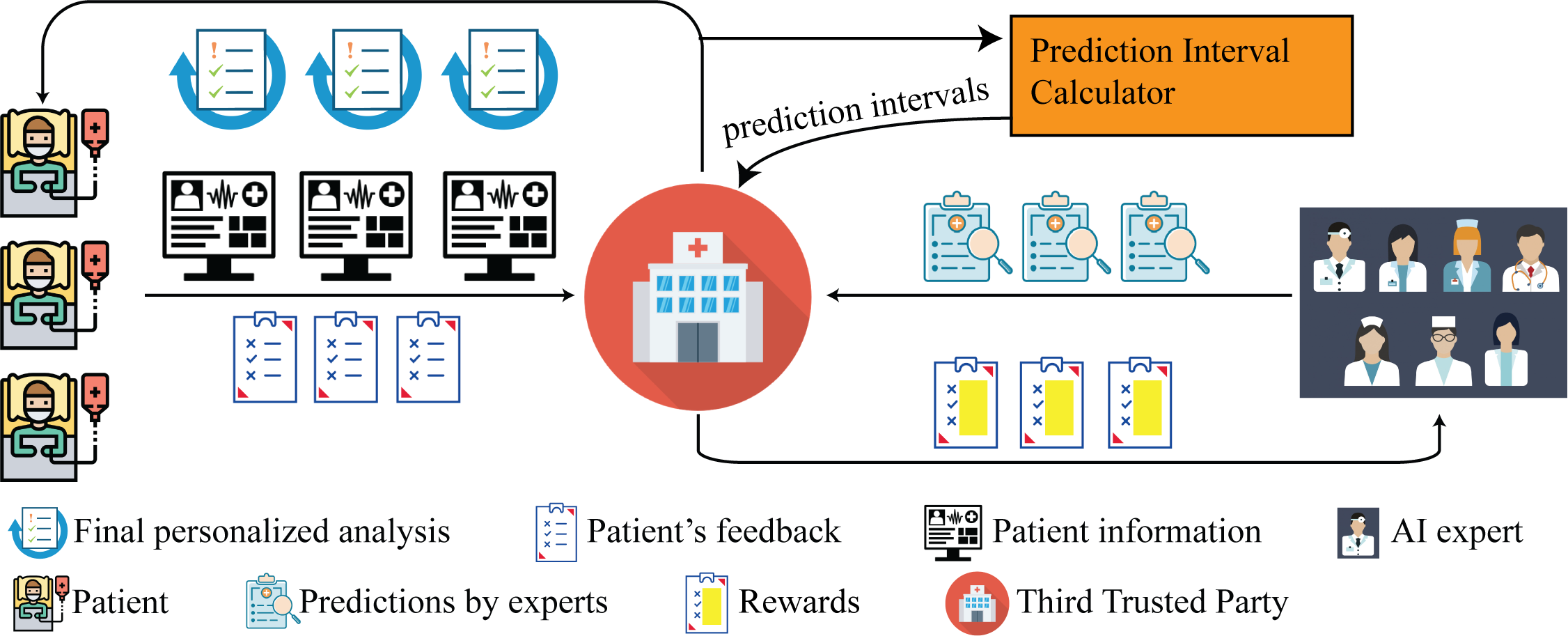}
    \caption{Workflows of Sepsyn-OLCP.}
    \label{fig:cpbandit_overview}
\end{figure*}

\begin{algorithm}[ht]
\caption{Sepsyn-OLCP.}
\label{algo:cp_bandit}
\SetKwInOut{Require}{Require}
\SetKwInOut{Output}{Output}
\Require{Offline training EHRs $\mathcal{D}_n = \{(x_{i,k}, r_{i,k})\}_{i=1}^T$; number of arms $K$; exploration budget $T$; candidate AI clinicians set $\mathcal{A}$; significance level $\alpha$; aggregation function $\phi$; number of bootstrap $B$; $refit\_step; IsRefit$}
\For{$t > T$}{
    \For{$k = 1, \dots, K$}{
        \tcp{Apply \ref{algo:EnsembleCP0} on historical data $\mathcal{D}_n$ to get LOO reward ensemble estimators and LOO residuals}
        $\hat{f}^{\phi,B}_{t, k} (x_{t, k}),\hat{e}_{t,k}^{\phi}$ = EnsembleCP0($\mathcal{D}_n,\mathcal{D}_n',k, A_k, \phi, B,refit\_step ,IsRefit$)\;

    }
    Compute the inverse empirical quantile function (ICDF) $F^{-1}_{t, k} (\alpha) := \alpha$ quantile of $\{ \hat{e}_{t,k}^{\phi} \}_{t=T}^{T+T_1}$ and \;
    $F^{-1}_{t, k} (1 - \alpha + \beta) := 1 - \alpha + \beta$ quantile of $\{ \hat{e}_{t,k}^{\phi}\}_{t=T}^{T+T_1}$\;
    
    Compute $\hat{\beta}_{t, k} := \arg \min_{\beta \in [0, \alpha]}\{ F^{-1}_{t, k} (1 - \alpha + \beta) - F^{-1}_{t, k} (\beta)\}$\;
    $U^{\alpha}_{t, k} (x_{t, k}) := \hat{f}^{\phi,B}_{t, k} (x_{t, k}) + F^{-1}_{t, k} (1 - \alpha + \hat{\beta}_{t, k})$\;
    $L^{\alpha}_{t, k} (x_{t, k}) := \hat{f}^{\phi,B}_{t, k} (x_{t, k}) + F^{-1}_{t, k} (\hat{\beta}_{t, k})$\;
    $B^{\alpha}_{t, k} (x_{t, k}) := \max_{a \neq k} U^{\alpha}_{t, a} (x_{t, a}) - L^{\alpha}_{t, k} (x_{t, k})$ for each $k = 1, \dots, K$\;
    $s^{\alpha}_{t, k} (x_{t, k}) := U^{\alpha}_{t, k} (x_{t, k}) - L^{\alpha}_{t, k} (x_{t, k})$\;
    $J_t := \arg \min_k B^{\alpha}_{t, k} (x_{t, k})$\;
    $j_t := \arg \max_k U^{\alpha}_{t, k} (x_{t, k})$\;
    
  \Output{Select $a_t := \arg \max_{k \in \{ j_t, J_t \}} s^{\alpha}_{t, k} (x_{t, k})$ and receive reward $r_{t, a_t}$\;}
}
\end{algorithm}

\begin{algorithm}[ht]
\caption{EnsembleCP0.
}
\label{algo:EnsembleCP0}
\SetKwInOut{Require}{Require}
\SetKwInOut{Output}{Output}
\Require{
Offline training EHRs $\mathcal{D}_n = \{(x_{i,k}, r_{i,k})\}_{i=1}^T$; 
EHRs of the current testing patient $\mathcal{D}_n' = \{(x_{j,k}, r_{j,k})| j>T\}$;  
prediction algorithm $A_k \in \mathcal{A}$, AI clinician index $k$;
aggregation function $\phi$;
number of bootstrap models $B$, $refit\_step$, $IsRefit$}

\If{$t = T + 1$}{
    \tcp{Initial fitting}
    \For{$b = 1, \dots, B$}{
        Sample with replacement an index set $S_b = (i_1, \dots, i_N)$ from indices $(1, \dots, N)$\;
        Compute $\hat{f}_{t,k}^b = A_k((x_{i,k}, r_{i,k}) | i \in S_b)$\;
        Save $B$ fitted estimators $\{\hat{f}_{t,k}^b\}|_{b=1}^B$\;
    }
}
\Else(\tcp*[h]{Re-fitting if required}){
\If{$IsRefit$==$True$}{    
   \If{$(t - T ) \mod refit\_step = 0$}{
    Retrain $\{\hat{f}_{t,k}^b\}_{b=1}^B$ on the updated dataset\;
    }}
    \Else{
    Load previous $B$ fitted estimators $\{\hat{f}_{t-1,k}^b\}|_{b=1}^B$ as $\{\hat{f}_{t,k}^b\}|_{b=1}^B$ to make predictions for the current patient\;
    }
}
\If{$t > T $}{
    $\hat{f}_{t,k}^{\phi,B}(x_{t,k}) = \phi(\{\hat{f}^b_{t,k} \}_{b=1}^B)$\;
    Compute $\hat{e}_{t,k}^{\phi} = r_{t,k} -\hat{f}_{t,k}^{\phi,B}(x_{t,k})$\;
}
\Output{$\hat{f}_{t,k}^{\phi,B}(x_{t,k})$, and $\hat{e}_{t,k}^{\phi}$}
\end{algorithm}
The proposed \textbf{Sepsyn-OLCP} algorithm operates within a \textbf{conformal prediction} and \textbf{gap-based Bayesian bandit} framework, specifically designed for clinical decision-making scenarios involving AI clinicians ($i.e., A_1, A_2, \cdots, A_K$ in Fig. \ref{fig:cpbandit_highlevel}). The algorithm is structured into two main phases: 1) apply EnsembleCP0 on historical data to get the LOO reward ensemble estimators and LOO residuals; 2) Update online selector to adjust the selection strategy. In this context, each arm corresponds to a distinct AI clinician, and each round represents the arrival of a new patient whose data is evaluated by the AI clinicians. A trusted third party, such as a hospital, handles the collection of rewards and computation of regrets to ensure unbiased and accurate performance evaluation. The workflows and key components of Sepsyn-OLCP are shown in \ref{fig:cpbandit_overview}. We denote $P_i$ as the patient at round $i$. Given the AI clinician $A_k$, each patient $P_i$ in the training dataset is associated with data $\{(x_{i,k}, r_{i,k})\}|_{i=1}^T$. Similarly, each patient $P_i$ in the testing dataset is associated with data $\{(x_{i,k}, r_{i,k})\}|_{i=T+1}^{T+T_1}$ given the AI clinician $A_k$.  

For each AI clinician (or arm) $A_k$ from $A_1$ to $A_K$, Sepsyn-OLCP evaluates the incoming patient data at each time step $t > T$. Sepsyn-OLCP leverages the \textbf{EnsembleCP0} method (Line 3), which applies LOO conformal prediction techniques to estimate ensemble rewards $\hat{f}^{\phi, B}_{t, k}(x_{t, k})$ and compute leave-one-out (LOO) residuals $\hat{e}_{t, k}^{\phi}$. These estimates form the basis for constructing \textbf{prediction intervals} that capture the uncertainty of the AI clinician’s performance.

The inverse empirical quantile function (ICDF) $F^{-1}_{t, k}(\alpha)$ is derived as the $\alpha$ quantile of residuals $\{\hat{e}_{t, k}^{\phi}\}_{t=T}^{T+T_1}$ (Line 5), while $F^{-1}_{t, k}(1 - \alpha + \beta)$ is calculated as the $1 - \alpha + \beta$ quantile (Line 5). The optimal parameter $\hat{\beta}_{t, k}$, minimizing the width of the prediction interval, is identified (Line 6). Using these intervals, Sepsyn-OLCP calculates the upper and lower confidence bounds, $U^{\alpha}_{t, k}(x_{t, k})$ and $L^{\alpha}_{t, k}(x_{t, k})$, respectively (Lines 7-8).

The \textbf{gap-based bandit} strategy is employed to guide decision-making. The gap $B_{t, k}(x_{t, k})$ quantifies the difference between the upper bound of an alternative AI clinician and the lower bound of the current AI clinician, emphasizing the need to exploit or explore other options (Line 9). The spread $s_{t, k}(x_{t, k})$ (Line 10) measures the uncertainty associated with each AI clinician's performance. The algorithm then identifies the AI clinician $J_t$ with the smallest gap (indicating a promising but uncertain option) and the clinician $j_t$ with the highest upper bound (suggesting the best current prediction) (Lines 11-12).

Ultimately, the expected best-performing AI clinician $a_t$ is chosen based on the maximum spread $s_{t, k}(x_{t, k})$ between these selected AI clinicians. The hospital, acting as the trusted third party, records the received reward $r_{t, a_t}$, processes the outcome, and calculates the regret associated with the decision to assess the long-term performance of the AI clinicians.

The \textbf{EnsembleCP0} subroutine uses conformal prediction principles to ensure robust and calibrated uncertainty estimates for each AI clinician. At the initial testing time step $t = T + 1$, $B$ bootstrap models are trained on resampled patient data to create a set of fitted estimators $\{\hat{f}_{t, k}^b\}_{b=1}^B$ (Lines 2-5). If re-fitting is required, the models are retrained periodically according to the \texttt{refit\_step} parameter, ensuring the estimators remain up-to-date with new patient data (Lines 6-9). If not, the previously fitted models are used to make predictions for the current patient (Lines 10-11).

For subsequent time steps $t > T$, the aggregated prediction $\hat{f}_{t, k}^{\phi, B}(x_{t, k})$ is computed using the aggregation function $\phi$, and the residual $\hat{e}_{t, k}^{\phi}$ is determined (Lines 13-14). The subroutine outputs the ensemble prediction and the associated residual, which are crucial for the conformal prediction intervals.

This workflow integrates conformal prediction to provide reliable uncertainty estimates and employs a gap-based Bayesian bandit approach for dynamic decision-making in a clinical setting, ensuring effective and interpretable patient care.

\subsection{Theoretical Analysis of Sepsyn-OLCP}
Our theoretical analysis is based on the top of \cite{EnbPI,2012simple,2014simple}. 
For each $t>T$, we define the event $\mathcal{E}_t:=\{L^{\alpha}_{t,k}\leq r_{t,k}\leq U^{\alpha}_{t,k} \ \forall k \in \{1,2,\cdots, K\}\}$ that ensures valid coverage of each stochastic reward at time $t$. For simplicity, we may remove the dependency on $x_{t,k}$ and remove the subscript $t$ when the time index is clear (e.g., $B_k=B_{t,k}(x_{t,k})$ at decision time $t$). Then, we have the following lemmas.

\begin{assumption}[Estimation quality \cite{EnbPI}]\label{lem:estimation_quality}
There exists a real sequence $\{\delta_T\}_{T\geq 1}$ such that
\[
\frac{1}{T} \sum_{t=1}^T \left( \hat{f}_{-t}(x_t) - f(x_t) \right)^2 \leq \delta_T^2 \]
and 
\[
\left| \hat{f}_{-(T+1)}(x_{T+1}) - f(x_{T+1}) \right| \leq \delta_T.
\]  
\end{assumption}
\begin{lemma}[Regret Bound]\label{lem:bound_1}
    Assume the event $\mathcal{E}_t$ occurs and $a_t\neq a_t^*$, which is the best arm at round $t$. Then
    \[
        R_{a_t}\leq B_{a_t},
    \]
    where $B_{a_t}:=\max_{k\neq a_t} U^{\alpha}_k-L^{\alpha}_{a_t}$
\end{lemma}
\textbf{Proof:}
We have on the event $\mathcal{E}_t$ that
\[
 R_{a_t} = r_{a_t^*}-r_{a_t} = \max_{k \neq a_t} r_k -r_{a_t}
    \leq \max_{k \neq a_t} U^{\alpha}_k-L^{\alpha}_{a_t} = B_{a_t}.
\]   
 

\begin{lemma}[Bound on contextual gap]\label{lem:bound_2}
    At time $t$, denote $s_k:=U^{\alpha}_k-L^{\alpha}_k$ as the width of the confidence interval. Then, 
    \[
    B_{a_t}\leq s_{a_t}.
    \]
\end{lemma}
\textbf{Proof:}
By the selection of the arm in Algorithm \ref{algo:cp_bandit}, we must have $a_t\in \{j_t,J_t\}$. We consider both cases.

Assume $a_t=j_t$. Then,
\[
B_{a_t}=B_{j_t}=\max_{k\neq j_t} U^{\alpha}_k-L^{\alpha}_{j_t}
    \leq U^{\alpha}_{j_t}-L^{\alpha}_{j_t}=s_{j_t}.
    \]

Assume $a_t=J_t$. We consider two cases based on the relationship of $s_{J_t}, s_{j_t}$. 

First, suppose $s_{J_t}<s_{j_t}$. Since $J_t$ is selected, this situation happens only when $j_t=J_t \Leftrightarrow U^{\alpha}_{j_t}=U^{\alpha}_{J_t}=\max_k U^{\alpha}_k$. Thus, 
\[
B_{a_t}=B_{J_t}=\max_{k\neq J_t} U^{\alpha}_k-L^{\alpha}_{J_t}
    \leq U^{\alpha}_{j_t}-L^{\alpha}_{J_t}=U^{\alpha}_{J_t}-L^{\alpha}_{J_t}=s_{J_t}.
\]

Next, suppose $s_{J_t}\geq s_{j_t}$. In this case, we must have $U^{\alpha}_{j_t}\leq U^{\alpha}_{J_t}$. Suppose not (i.e., $U^{\alpha}_{j_t}> U^{\alpha}_{J_t}$), then $L^{\alpha}_{j_t}> L^{\alpha}_{J_t}$. As a result,
\[
    B_{j_t} = \max_{k\neq j_t} U^{\alpha}_k-L^{\alpha}_{j_t} < U^{\alpha}_{j_t}-L^{\alpha}_{J_t} = B_{J_t}.
\]
However, $J_t:=\arg\min_k B_k$, so that this is a contradiction. As a consequence, 
\begin{align*}
    B_{a_t}=B_{J_t}&=\max_{k\neq J_t} U^{\alpha}_k-L^{\alpha}_{J_t}\\
    &\leq U^{\alpha}_{j_t}-L^{\alpha}_{J_t} \\
    & \leq U^{\alpha}_{J_t}-L^{\alpha}_{J_t}=s_{J_t}.
\end{align*}

\begin{lemma}\label{lem:prob_1}
For any time $t$ and any arm $k$, recall $N_{t,k}$ is the number of times that arm $k$ is pulled by round $t$. 
Assume that 
(1) the errors $\{\epsilon_{\tau,k}\}_{\tau=1}^{N_{t,k}}$ are independent and identically distributed (i.i.d.) according to a common CDF $F_{t,k}$, which is Lipschitz continuous with constant $L_{t,k}>0$; and (2) there is a real sequence $\{\delta_{T,k}\}_{T\geq 1}$ that converges
to zero such that $\sum_{\tau=1}^{N_{t,k}} (\hat{f}^{\phi}_{-\tau,k}(x_{\tau,k})-f_k(x_{\tau,k}))^2/N_{t,k} \leq \delta_{N_{t,k},k}^2$. Then, 
\[
\mathbb{P}(\mathcal{E}_t)\geq 1-\sum_{k=1}^K\left(\alpha + 24 \sqrt{\log (16 N_{t,k}) / N_{t,k}}+4 L_{t,k} \delta_{N_{t,k},k}^{2 / 3}\right).
\]
\end{lemma}

\begin{lemma}[Bound of confidence interval width]\footnote{Based on Theorem 3 in \cite{EnbPI}}\label{lem:bound_3}
Suppose assumptions in Lemma \ref{lem:prob_1} hold. In addition, assume there exists a sequence $\{\gamma_T\}_{T\geq 1}$ that converges to zero such that $|\hat{f}^{\phi}_{-t,k}(x_{t,k})-f_k(x_{t,k})|\leq \gamma_{N_{t,k}}$. Lastly, assume that $F^{-1}_{t,k},\hat{F}^{-1}_{t,k},\hat{F}_{t,k}$ are Lipschitz continuous with constants $K_{t,k},K'_{t,k},K^{''}_{t,k}$ respectively. Thus,
\[
    |s_k-W^*_{t,k}(\alpha)| \leq C^*_{t,k}(\alpha)\left(\gamma_{N_{t,k}}+\sqrt{\log (16 N_{t,k}) / N_{t,k}}+\delta_{N_{t,k},k}^{2 / 3}\right ),
\]
where $W^*_{t,k}(\alpha):=\min_{\beta \in [0,\alpha]} \hat{F}^{-1}_{t,k}(1-\alpha+\beta)-\hat{F}^{-1}_{t,k}(\beta)$ denotes the fixed oracle interval width that solely depends on $\alpha$ and the constant $C^*_{t,k}(\alpha)$ that is a function of the Lipschitz constants and $\alpha$.
\end{lemma}

As a consequence of earlier lemmas, we can provide the following guarantee.
\begin{theorem}[Bound on simple regret]
\label{thm:bound_simple_regret}
Suppose the assumptions in Lemmas \ref{lem:bound_1}---\ref{lem:bound_3} hold and we build the confidence intervals for all arms at level $\alpha$. Fix $\epsilon>0$. Suppose $N_{t,a_t}$ is large enough so that \[C^*_{t,a_t}(\alpha)\left(\gamma_{N_{t,a_t}}+\sqrt{\log (16 N_{t,a_t}) / N_{t,a_t}}+\delta_{N_{t,a_t},a_t}^{2 / 3}\right )\leq \epsilon.\] Then,
\[
    \mathbb{P}(R_{a_t}\leq W^*_{t,a_t}(\alpha)+\epsilon)\geq \sum_{k=1}^K\left(\alpha + 24 \sqrt{\log (16 N_{t,k}) / N_{t,k}}+4 L_{t,k} \delta_{N_{t,k},k}^{2 / 3}\right).
\]
\end{theorem}

\textbf{Proof:}

The result easily follows by earlier Lemmas, where $R_{a_t}\leq B_{a_t} \leq s_{a_t}$ with probability at least $1-\mathbb{P}(\mathcal{E}_t)$. Because $s_{a_t}\rightarrow W^*_{t,k}(\alpha)$ as $N_{t,k}\rightarrow \infty$, after sufficiently many pulling of arm $k$, the deviation from $W^*_{t,a_t}(\alpha)$ is unlikely to be greater than $\epsilon$ by $\mathbb{P}(\mathcal{E}_t)$.

Note that the factor $W^*_{t,a_t}(\alpha)$ in \ref{thm:bound_simple_regret} occurs naturally. 
Consider an example when the simple regret becomes the difference of the largest and the $j$-th order statistics of errors (e.g., $R_{a_t}=\epsilon_{a_t^*}-\epsilon_{a_t}$). In this case, $W^*_{t,a_t}(\alpha)=W^*(\alpha):=\min_{\beta \in [0,\alpha]} \hat{F}^{-1}(1-\alpha+\beta)-\hat{F}^{-1}(\beta)$ is the smallest $(1-\alpha)$ confidence interval for any $\epsilon_k$. 
Because errors are i.i.d., $\mathbb{P}(\epsilon_k \in [L^{\alpha}_{k},U^{\alpha}_k]) \approx 1-\alpha$ for any arm $k$. Hence, $\mathbb{P}(R_{a_t}\geq W^*_{t,a_t}(\alpha)+\epsilon)\leq \mathbb{P}(\epsilon_{a_t^*}>U^{\alpha}_k)+\mathbb{P}(\epsilon_{a_t}<L^{\alpha}_k),$ where the latter is approximately bounded by $K\alpha$.

\begin{figure*}[htb!]
     \centering
     \begin{subfigure}{0.42\textwidth}
         \centering
         \includegraphics[width=\linewidth]{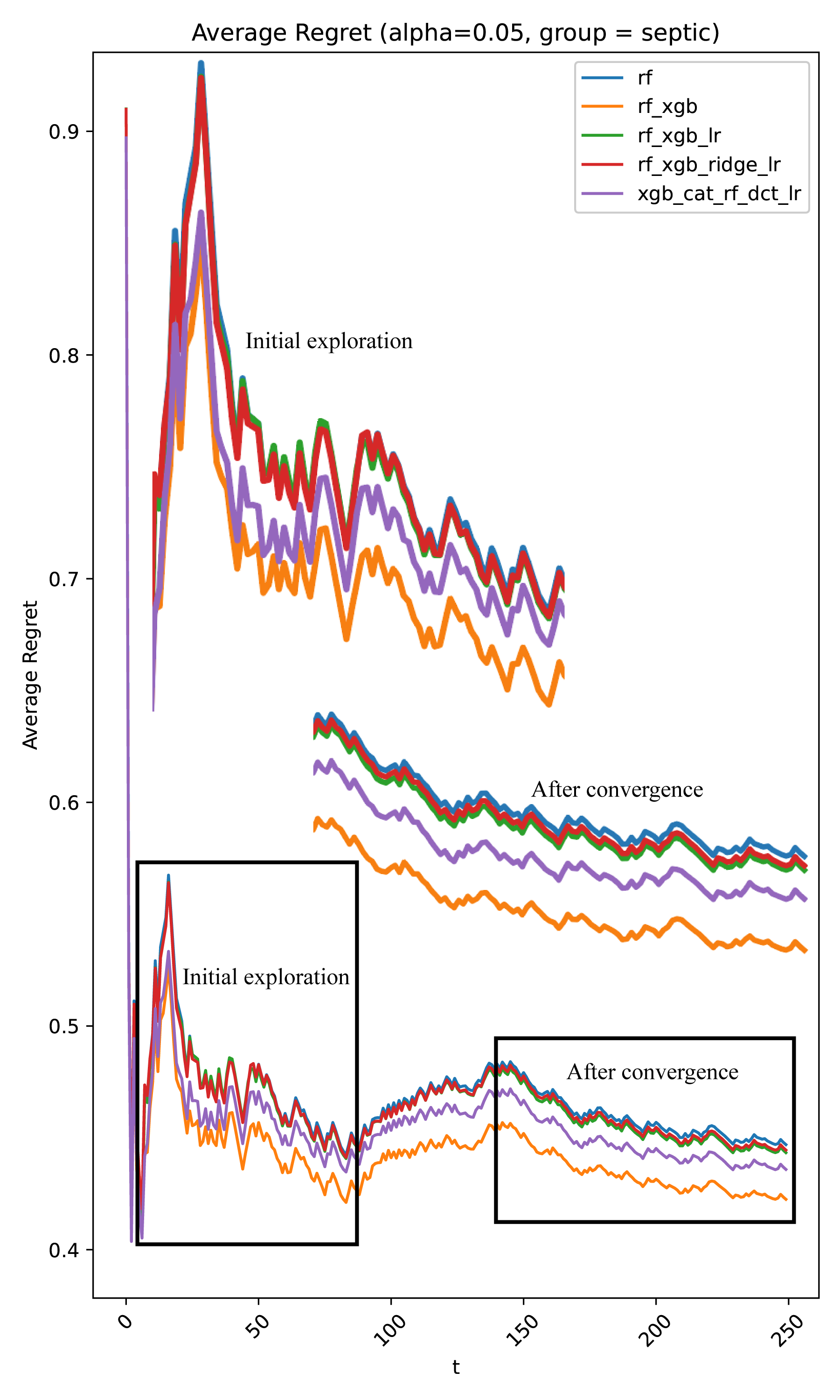}
         \caption{Average Regret ($\alpha = 0.05$).}
         \label{fig:AR_rfcombo005}
     \end{subfigure}
     \begin{subfigure}{0.4\textwidth}
         \centering
         \includegraphics[width=\linewidth]{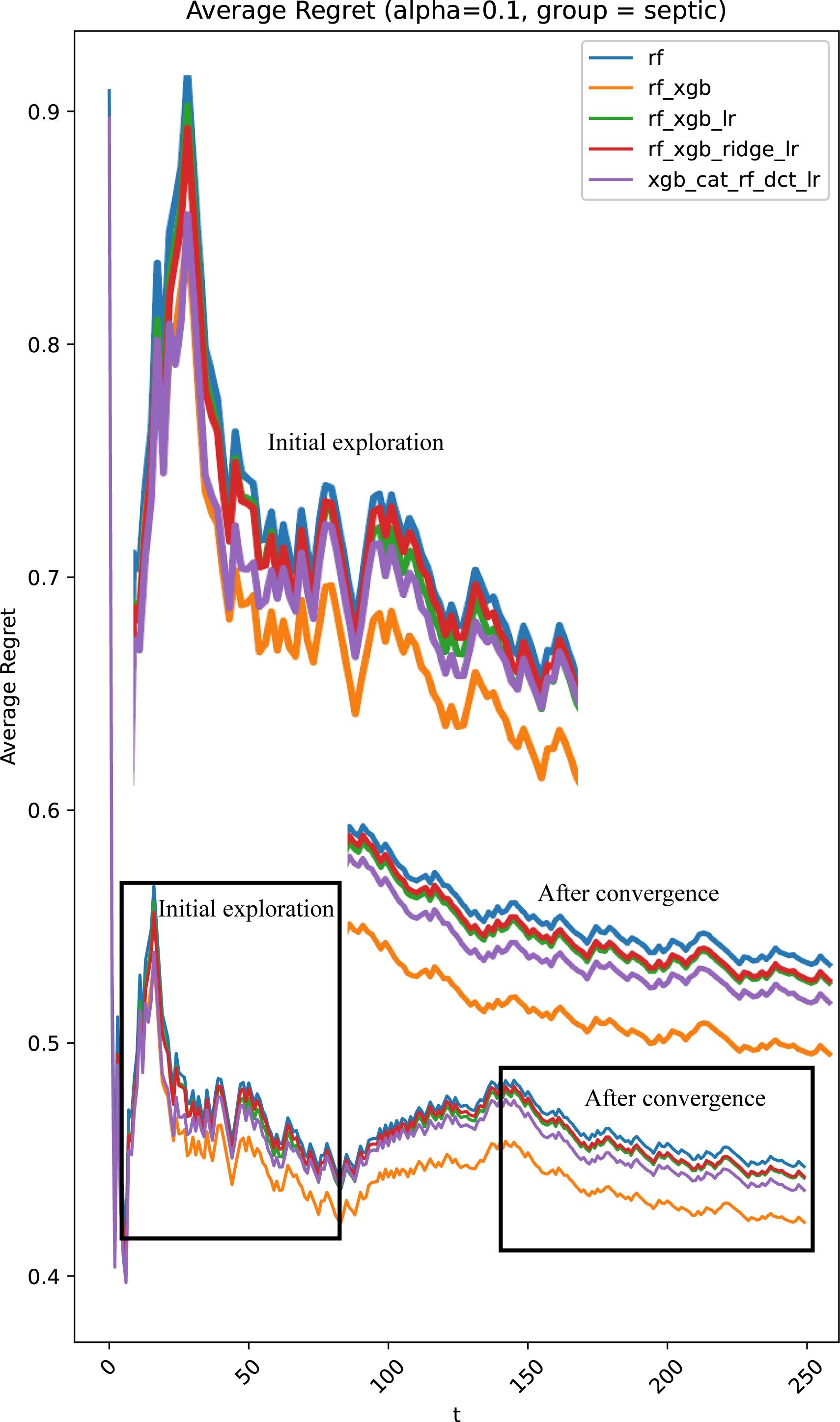}
         \caption{Average Regret ($\alpha = 0.1$)}
         \label{fig:AR_rfcombo01}
     \end{subfigure}
     \caption{Average Regret (baseline = rf).}
\end{figure*}

\subsection{Dataset Description and Preprocessing}
We use the 2019 PhysioNet Computing Cardiology Challenge \cite{physionetChallenge} dataset following the Sepsis-3 guidelines \cite{sepsis3}.  
The dataset contains EMRs of ICU patients from three different hospitals. The EMRs of each patient are stored in a .psv file on an hourly basis. Implementation of \ref{algo:EnsembleCP0} is based on \cite{EnbPI}, and we have rewritten \ref{algo:EnsembleCP0} in parallel computing mode. 
We utilized \textit{\textbf{HyperImpute}} \cite{jarrett2022hyperimpute} for data imputation, which leverages generalized iterative imputation enhanced with automatic model selection, to efficiently handle missing data across datasets, ensuring robust and adaptive imputation performance.

\section{Results}
\label{sec:res}

\subsection{Experimental setup}
We evaluate Sepsyn-OLCP using different $\alpha$ values, which control the confidence level in the conformal prediction framework. We have a list of AI clinicians as candidates, i.e., Neural Network (nnet), Random Forest (rf), XGBoost (xgb), Ridge Regression (ridge), Logistic Regression (lr), and Decision Tree (dct). All the AI clinicians are established using standard Python libraries instead of advanced algorithms to see the performance of Sepsyn-OLCP. If the average regret can decrease even when we add some ``dumb'' AI clinicians into the candidates, it means that Sepsyn-OLCP can effectively improve the performance through the online learning process compared with a standalone AI clinician. 

Different combinations of AI clinician models are plotted to observe the average regret over time. 
We use one AI clinician as the baseline for each combination. Each time, we add one AI clinician to the candidates to observe the changes in the average regrets of different combinations. 
 \begin{figure*}[htb!]
     \centering
     \begin{subfigure}{0.4\textwidth}
         \centering
         \includegraphics[width=\linewidth]{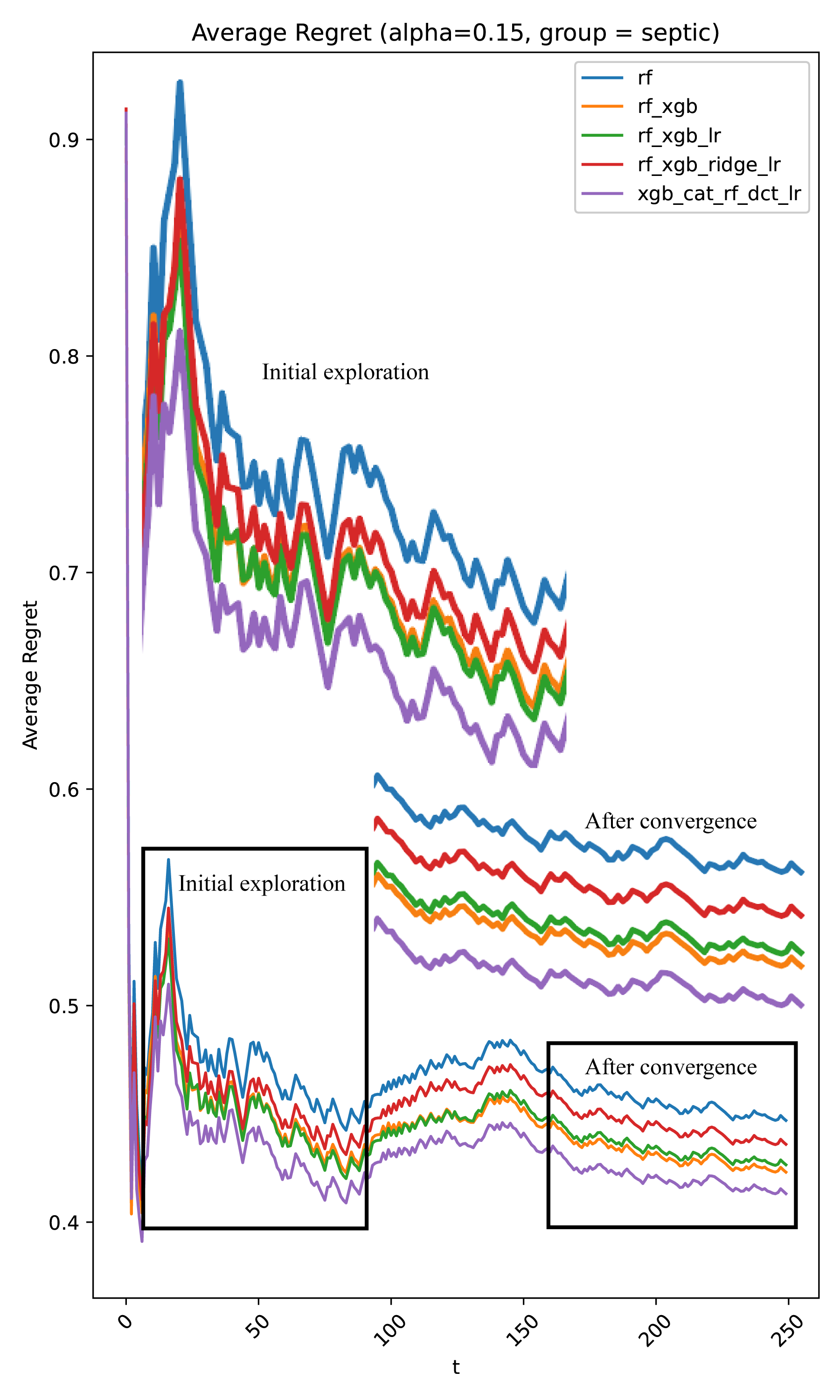}
         \caption{Average Regret ($\alpha = 0.15$).}
         \label{fig:AR_rfcombo015}
     \end{subfigure}
     \begin{subfigure}{0.4\textwidth}
         \centering
         \includegraphics[width=\linewidth]{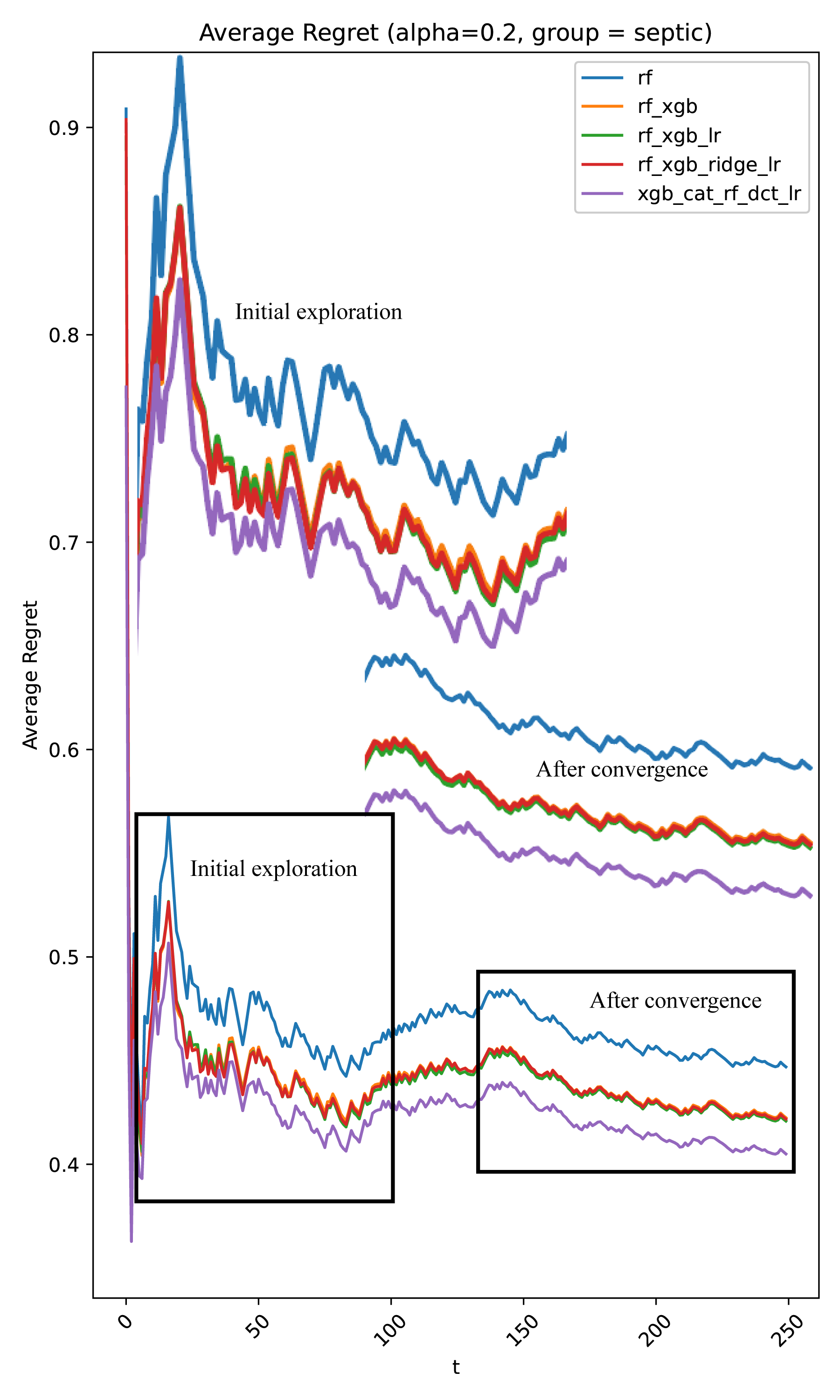}
         \caption{Average Regret ($\alpha = 0.2$).}
         \label{fig:AR_rfcombo02}
     \end{subfigure}
     \caption{Average Regret (baseline = rf).}
\end{figure*}
We randomly selected 1,000 septic patients and 1,000 non-septic patients for a balanced training dataset. For the testing dataset, we randomly selected 250 septic patients and 250 non-septic patients for all the experiments. We plot the average regret for the septic patients, and the performance in terms of AUROC, AUPRC, Accuracy, F-measure and Utility is calculated following the standards provided by the 2019 PhysioNet Computing Cardiology Challenge \cite{physionetChallenge}.
For implication details, please refer to our open-sourced code\footnote{\url{https://github.com/Annie983284450-1/CPGapBandit.git}}. All the experiments in this paper were performed using the resources provided by the \textbf{\textit{Partnership for an Advanced Computing Environment (PACE)}} \cite{PACE} at the Georgia Institute of Technology. 

 \begin{figure*}[htb!]
     \centering
     \begin{subfigure}{0.4\textwidth}
         \centering
         \includegraphics[width=\linewidth]{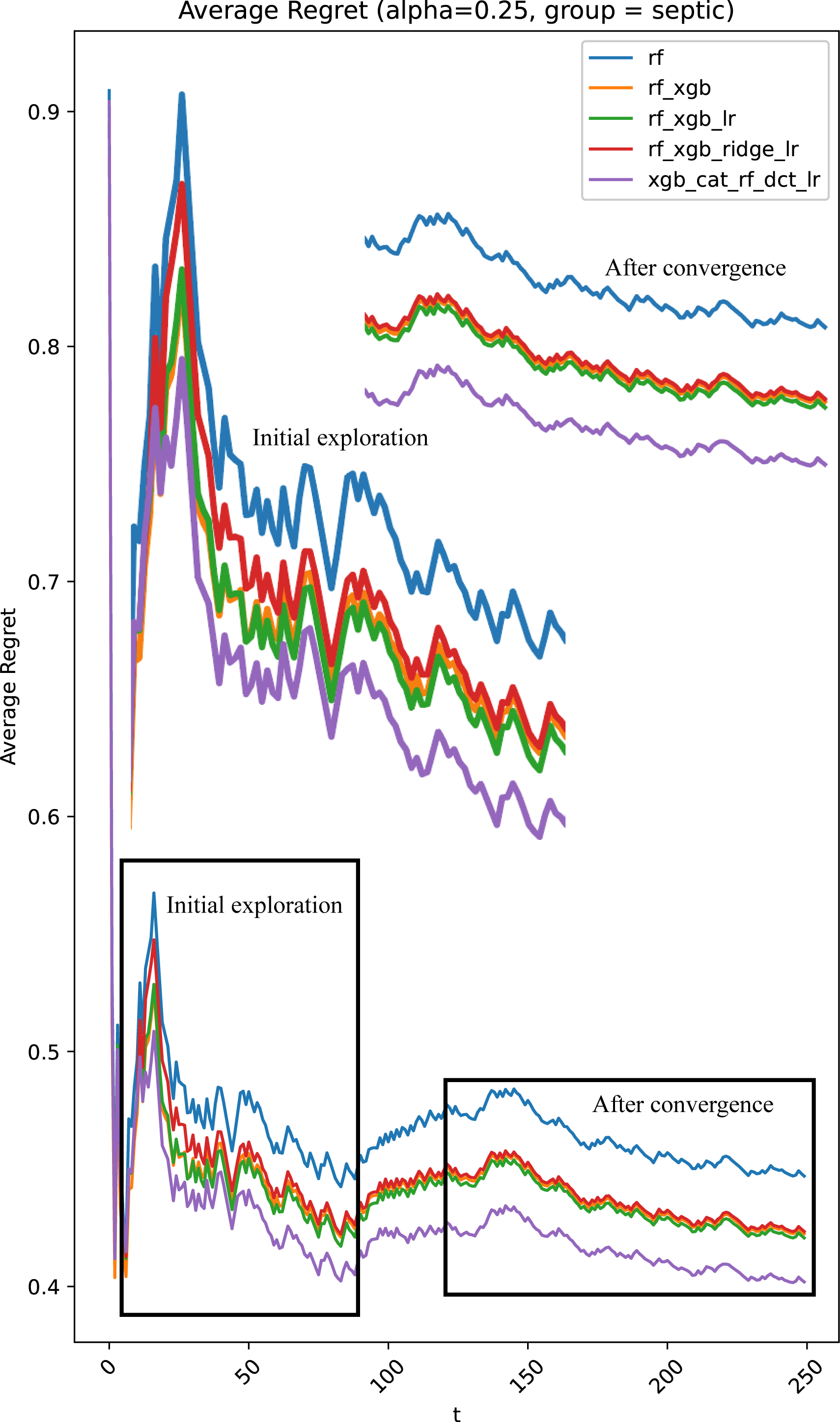}
         \caption{Average Regret ($\alpha = 0.25$, baseline = rf).}
         \label{fig:AR_rfcombo025}
     \end{subfigure}
     \begin{subfigure}{0.42\textwidth}
         \centering
         \includegraphics[width=\linewidth]{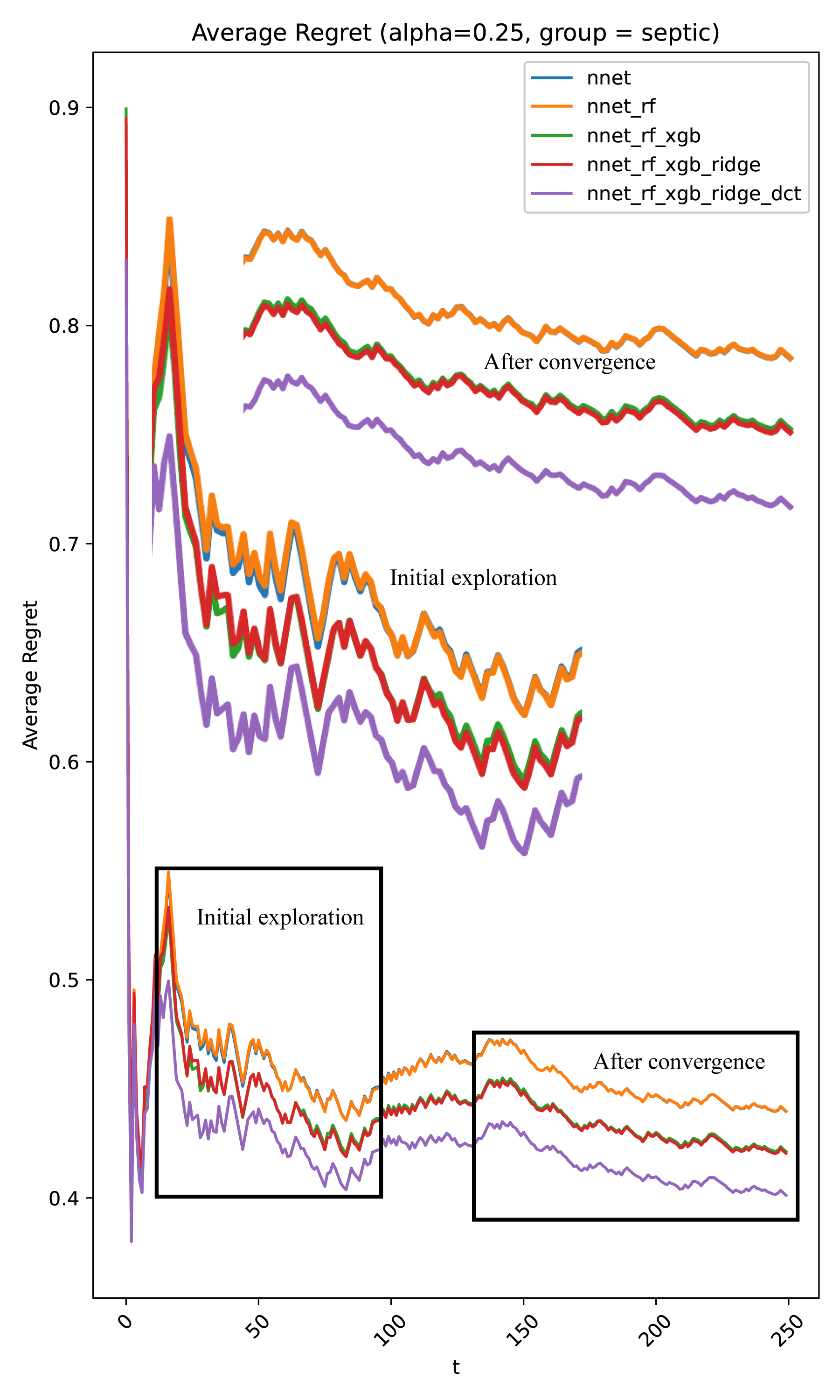}
         \caption{Average Regret ($\alpha = 0.25$, baseline = nnet).}
         \label{fig:AR_nnetcombo025}
     \end{subfigure}
     \caption{Average Regret (alpha = 0.25).}
\end{figure*}

\begin{figure*}[htb!]
    \centering
         \begin{subfigure}{0.4\textwidth}
         \centering
         \includegraphics[width=\linewidth]{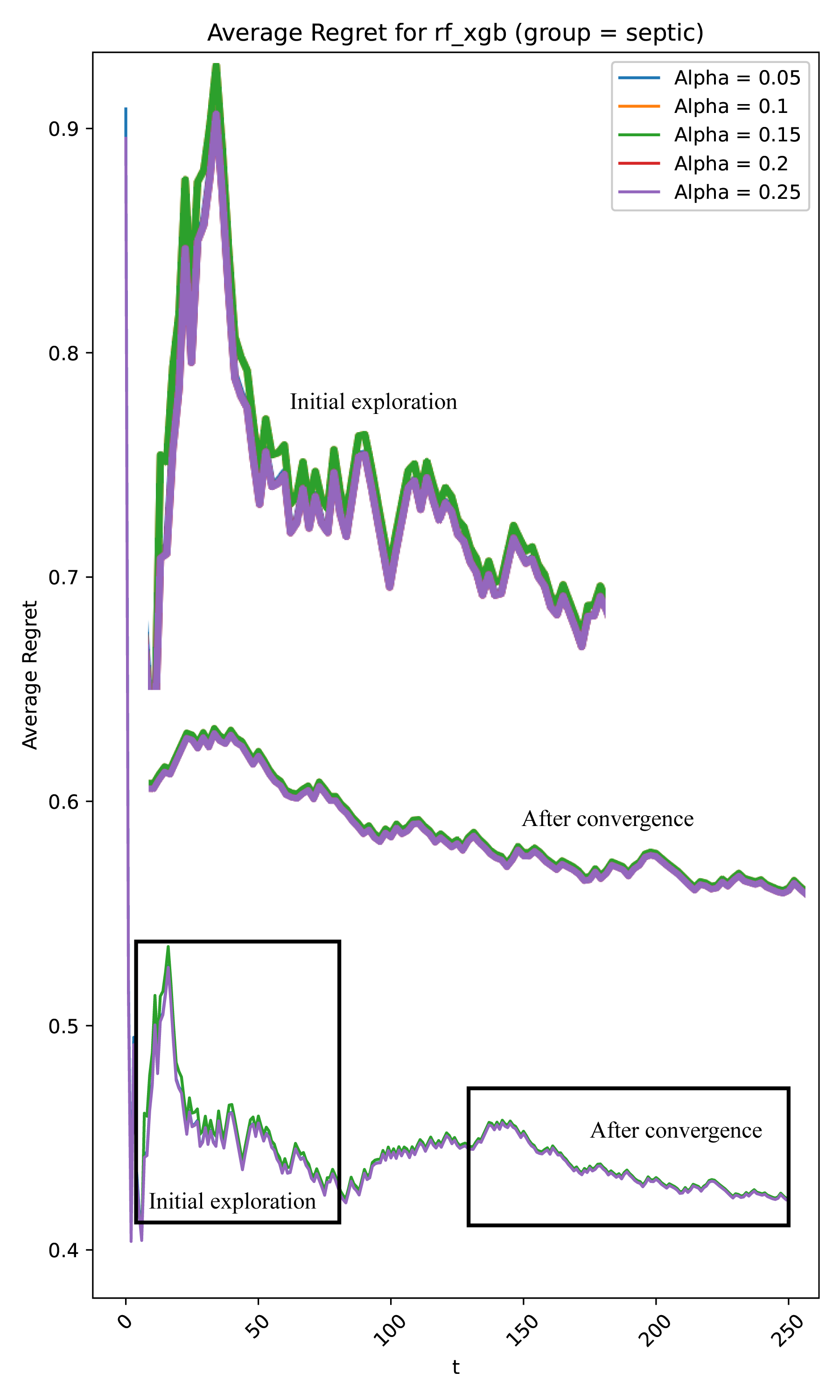}
          \caption{Average Regret (rf\_xgb).}
          \label{fig:AR_rf_xgb_differnt_alphas}
     \end{subfigure}
     \begin{subfigure}{0.4\textwidth}
         \centering
        \includegraphics[width=\linewidth]{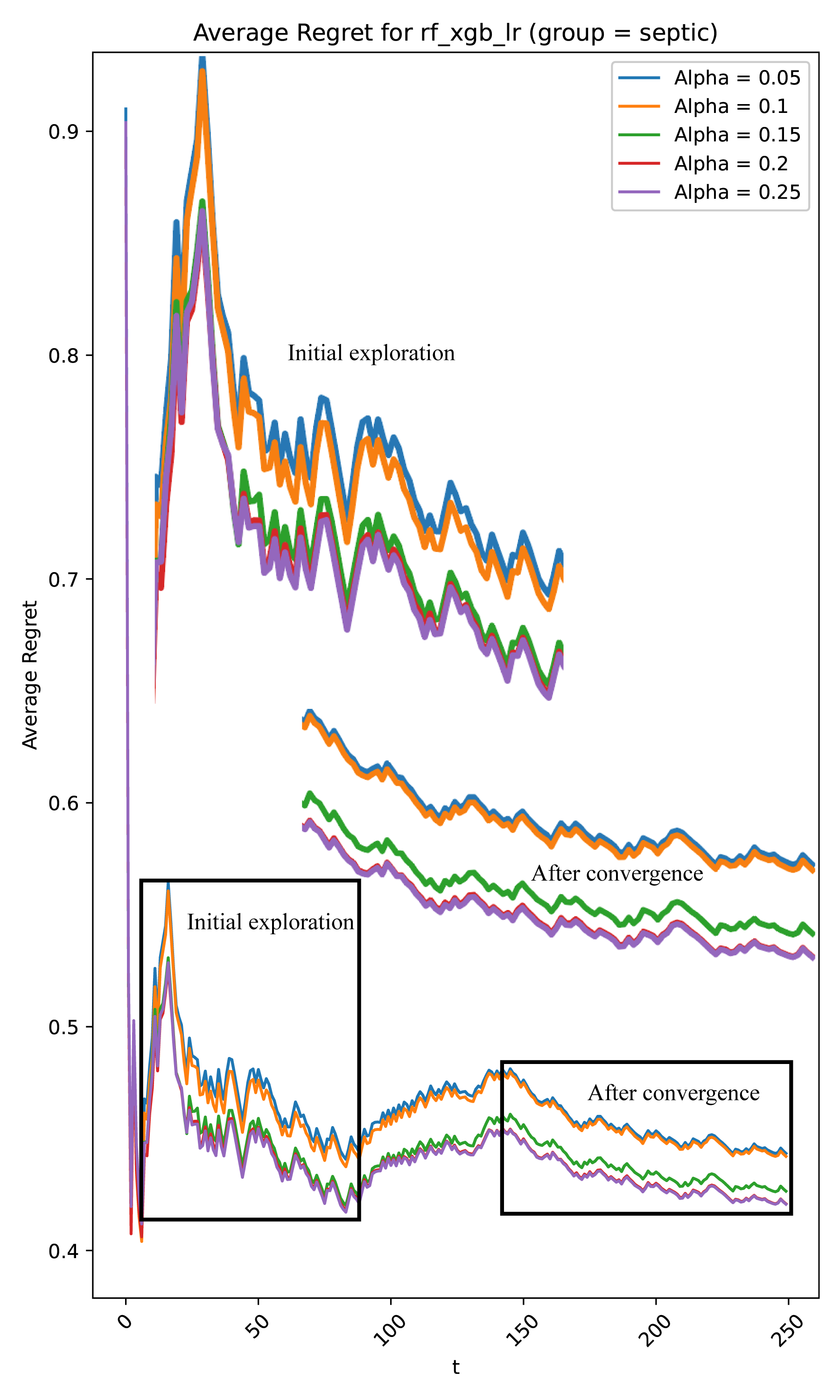}
         \caption{Average Regret (rf\_xgb\_lr).}
        \label{fig:AR_rf_xgb_lr_differnt_alphas}
     \end{subfigure}
      \caption{Average Regret of Different $\alpha$ (baseline = rf).}
\end{figure*}

\begin{figure*}[htb!]
    \centering
         \begin{subfigure}{0.4\textwidth}
         \centering
         \includegraphics[width=\linewidth]{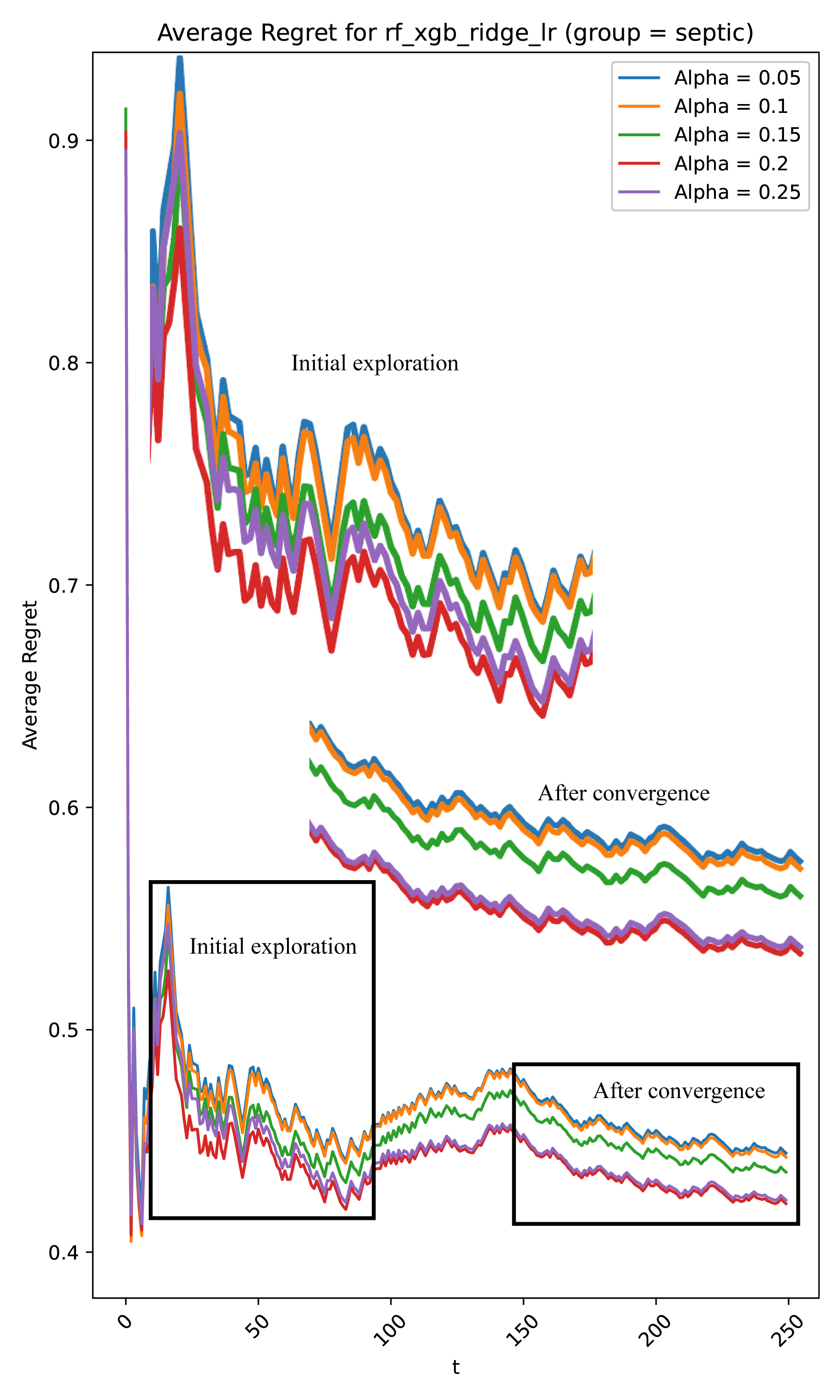}
           \caption{Average Regret (rf\_xgb\_ridge\_lr).}
         \label{fig:AR_rf_xgb_ridge_lr_differnt_alphas}
     \end{subfigure}
     \begin{subfigure}{0.4\textwidth}
         \centering
        \includegraphics[width=\linewidth]{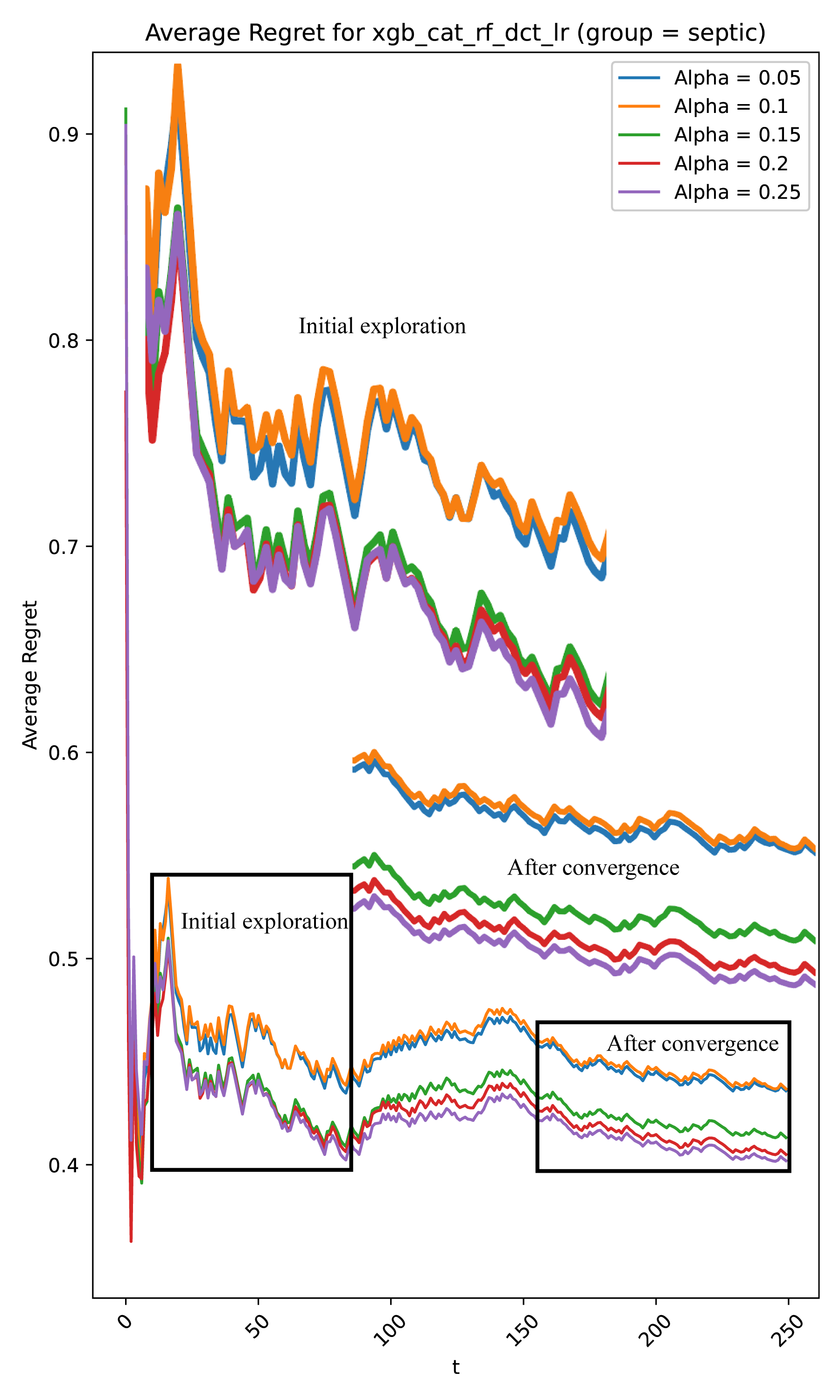}
         \caption{Average Regret (xgb\_cat\_rf\_dct\_lr).}
        \label{fig:AR_xgb_cat_rf_dct_lr_differnt_alphas}
     \end{subfigure}
          \caption{Average Regret of Different $\alpha$ (baseline = rf).}
\end{figure*}
\subsection{Effects of significance level $\alpha$ on Confidence Intervals and Online Exploration}
\begin{figure*}[htb!]
     \centering
     \begin{subfigure}{0.4\textwidth}
         \centering
         \includegraphics[width=\linewidth]{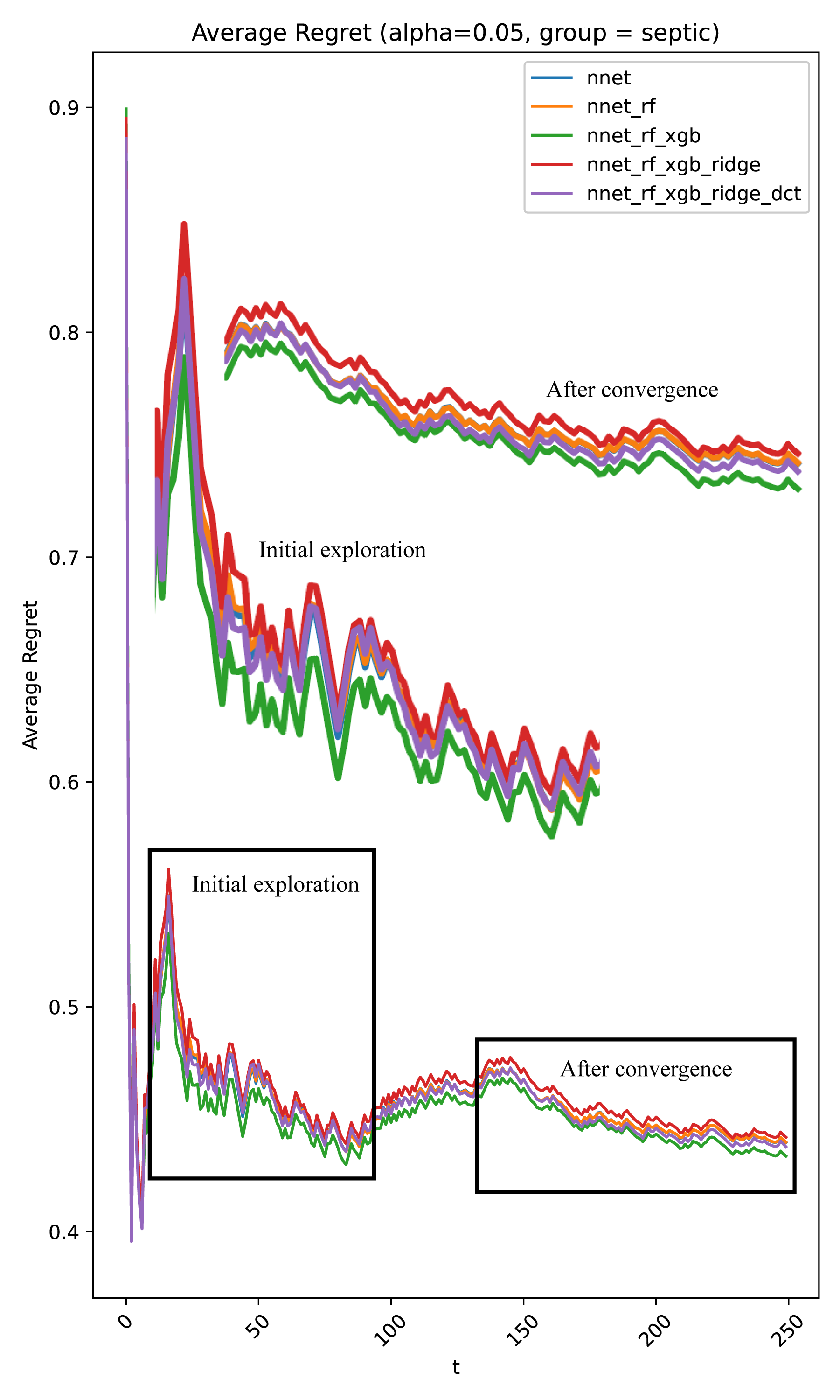}
         \caption{Average Regret ($\alpha = 0.05$).}
         \label{fig:AR_nnetcombo005}
     \end{subfigure}
     \begin{subfigure}{0.4\textwidth}
         \centering
         \includegraphics[width=\linewidth]{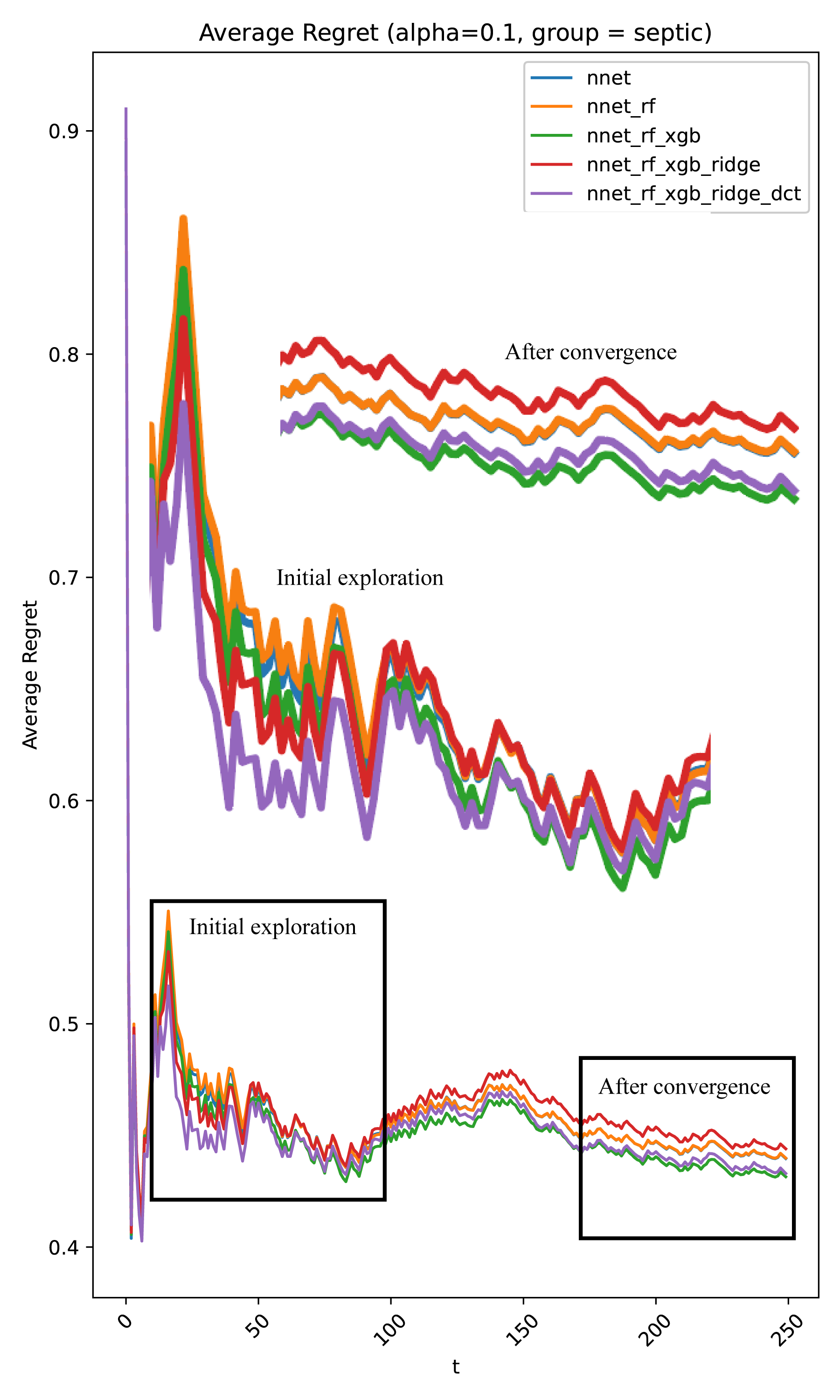}
         \caption{Average Regret ($\alpha = 0.1$).}
         \label{fig:AR_nnetcombo01}
     \end{subfigure}
     \caption{Average Regret (baseline = nnet).}
\end{figure*}

The significance level $\alpha$ controls the width of the confidence interval used in Sepsyn-OLCP. Lower values of $\alpha$ (e.g., $\alpha = 0.05$ and $\alpha = 0.1$) result in narrower confidence intervals, leading to more conservative exploration. When Sepsyn-OLCP makes conservative explorations with low $\alpha$ values, the algorithm relies heavily on existing knowledge and explores less. This behavior can be advantageous when the AI clinician candidates' predictions are already reasonably accurate, as it might reduce unnecessary exploration that could possibly increase regret. For example, as we can see in \ref{fig:AR_rfcombo005} and \ref{fig:AR_rfcombo01}, rf\_xgb combination has the lowest regrets compared with other combinations when the significance level $\alpha$ is low.   

\ref{fig:AR_rf_xgb_differnt_alphas}, \ref{fig:AR_rf_xgb_lr_differnt_alphas}, \ref{fig:AR_rf_xgb_ridge_lr_differnt_alphas}, and \ref{fig:AR_xgb_cat_rf_dct_lr_differnt_alphas} illustrate the average regret over time for various combinations of AI clinicians with different values of the parameter $\alpha$, specifically for the septic patients. Each figure presents the performance of a combination of experts across five different 
$\alpha$ values: 0.05, 0.1, 0.15, 0.2, and 0.25.  

All figures (i.e., \ref{fig:AR_rf_xgb_differnt_alphas}, \ref{fig:AR_rf_xgb_lr_differnt_alphas}, \ref{fig:AR_rf_xgb_ridge_lr_differnt_alphas}, and \ref{fig:AR_xgb_cat_rf_dct_lr_differnt_alphas}) show a relatively large average regret in the beginning, which gradually decreases as the exploration phase stabilizes. Generally, during the initial phase, the algorithm tries out various actions to learn about the environment and gather data to make more informed decisions in the future.

As time progresses, we can see that the average regrets decrease and converge, indicating that the algorithm has learned an effective policy and is optimizing its decisions. The separation of regret values across different 
$\alpha$ levels become more apparent.


\begin{figure*}[h]
     \centering
     \begin{subfigure}{0.4\textwidth}
         \centering
         \includegraphics[width=\linewidth]{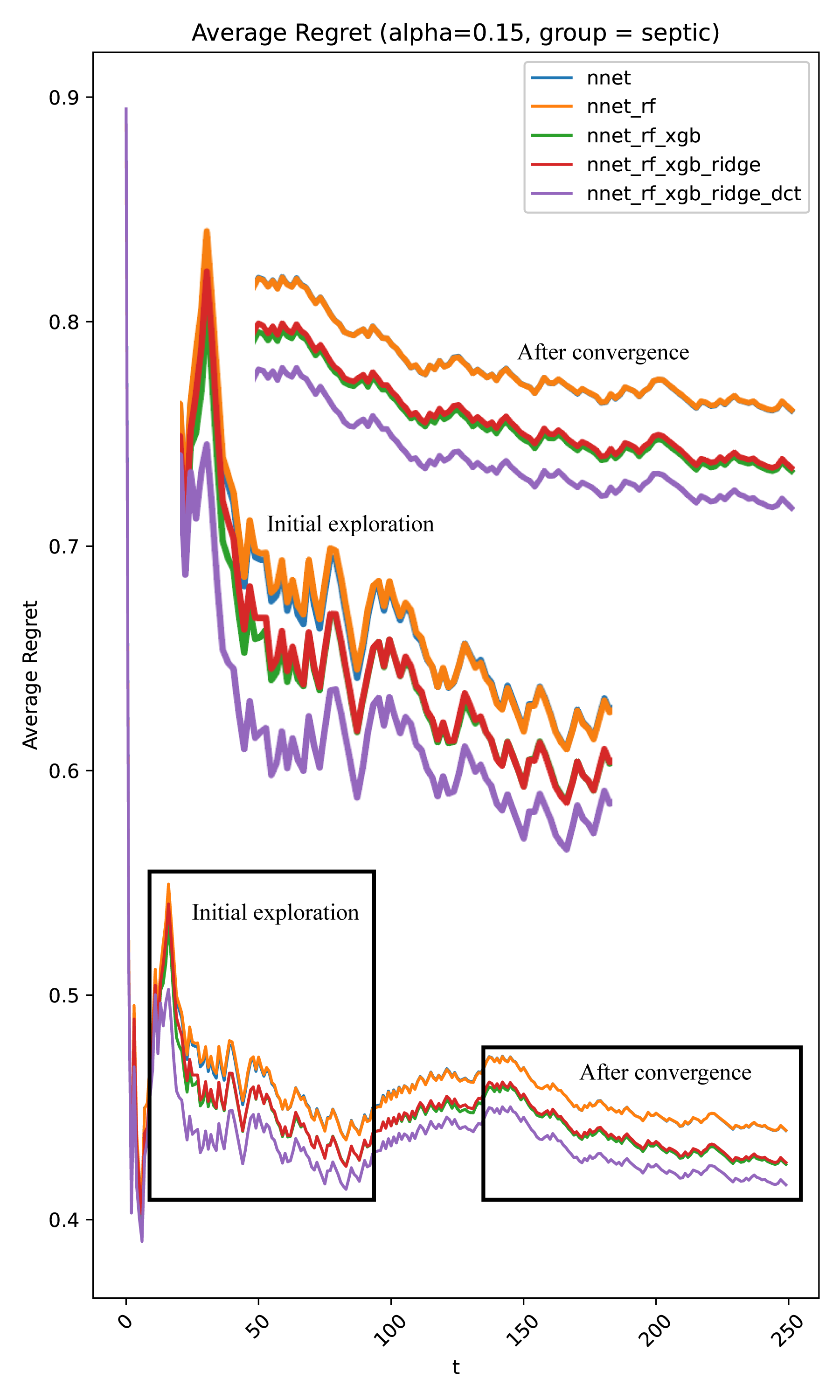}
         \caption{Average Regret ($alpha = 0.15$).}
         \label{fig:AR_nnetcombo015}
     \end{subfigure}
     \begin{subfigure}{0.4\textwidth}
         \centering
         \includegraphics[width=\linewidth]{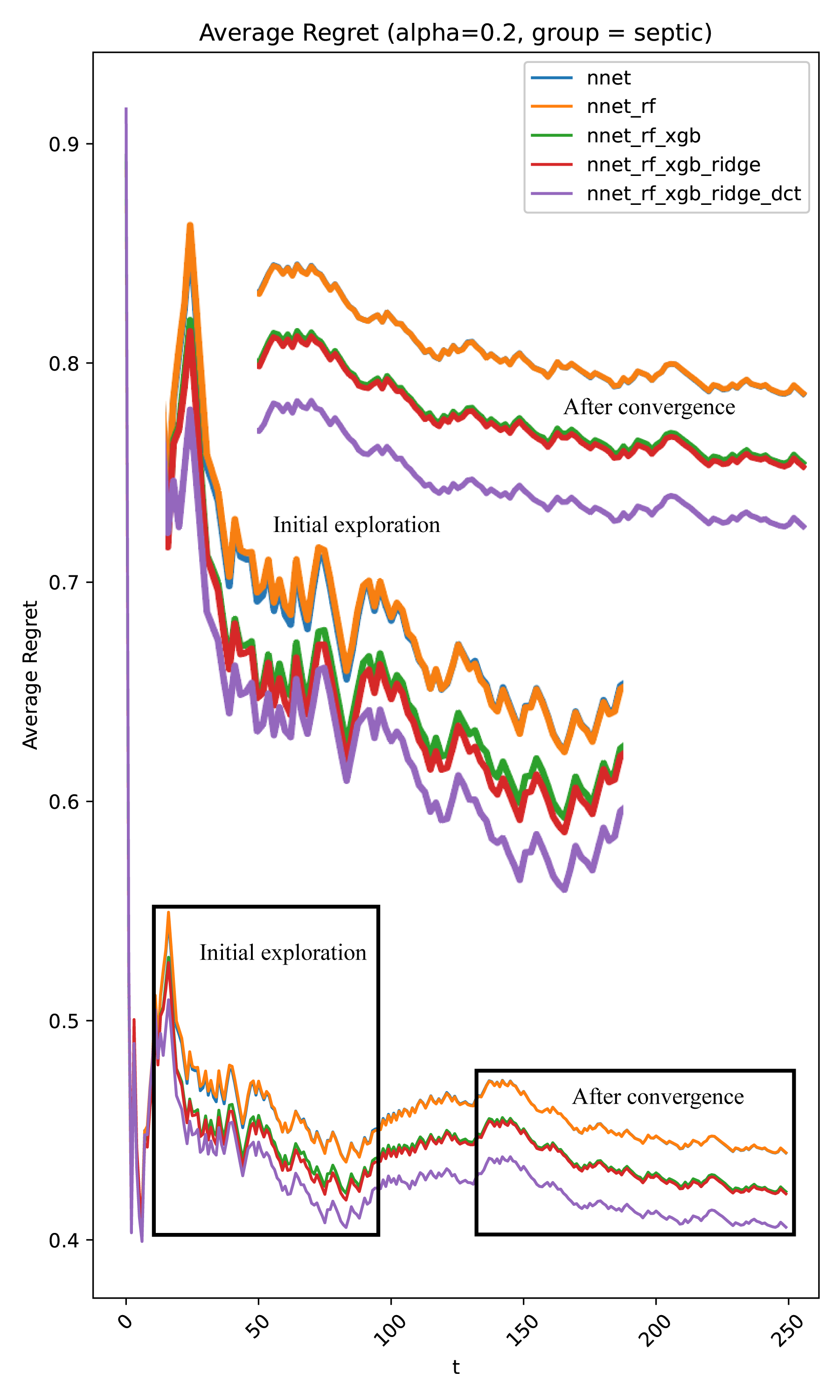}
         \caption{Average Regret ($\alpha = 0.2$).}
         \label{fig:AR_nnetcombo02}
     \end{subfigure}
     \caption{Average Regret (baseline = nnet).}
\end{figure*}
Lower values of $\alpha$ tend to show more consistent and higher regret in the long run. This suggests that lower values of $\alpha$ will result in more conservative exploration.
Higher values of $\alpha$ exhibit greater fluctuations in the early rounds and lower average regret in the long run. This is likely due to more aggressive exploration, leading to suboptimal choices in some cases in the early stages. 
The xgb\_cat\_rf\_dct\_lr combination involving all experts (i.e., XGBoost, CatBoost, Random Forest, Decision Tree, and Logistic Regression) shows the most variation across different 
$\alpha$ values, with significant performance differences between lower and higher 
$\alpha$.
The rf\_xgb\_ridge\_lr combination demonstrates a more balanced performance, with the average regret curves gradually decreasing and showing smaller gaps between different 
$\alpha$ levels.
The simpler combinations like rf\_xgb\_lr and rf\_xgb combinations result in smoother curves, with regret levels more tightly clustered together, particularly in the later stages.

\subsection{Complementary Strengths of Random Forest and XGBoost Combination}

Random Forest is a robust ensemble model that performs well on diverse datasets by averaging predictions from multiple decision trees, which helps mitigate overfitting and provides stable performance.
XGBoost is a powerful gradient boosting algorithm known for its ability to capture complex relationships in data through iterative, boosting-based learning. It tends to perform well when the data has subtle nonlinear patterns.

\begin{figure*}[h]
     \centering
     \begin{subfigure}{0.4\textwidth}
         \centering
         \includegraphics[width=\linewidth]{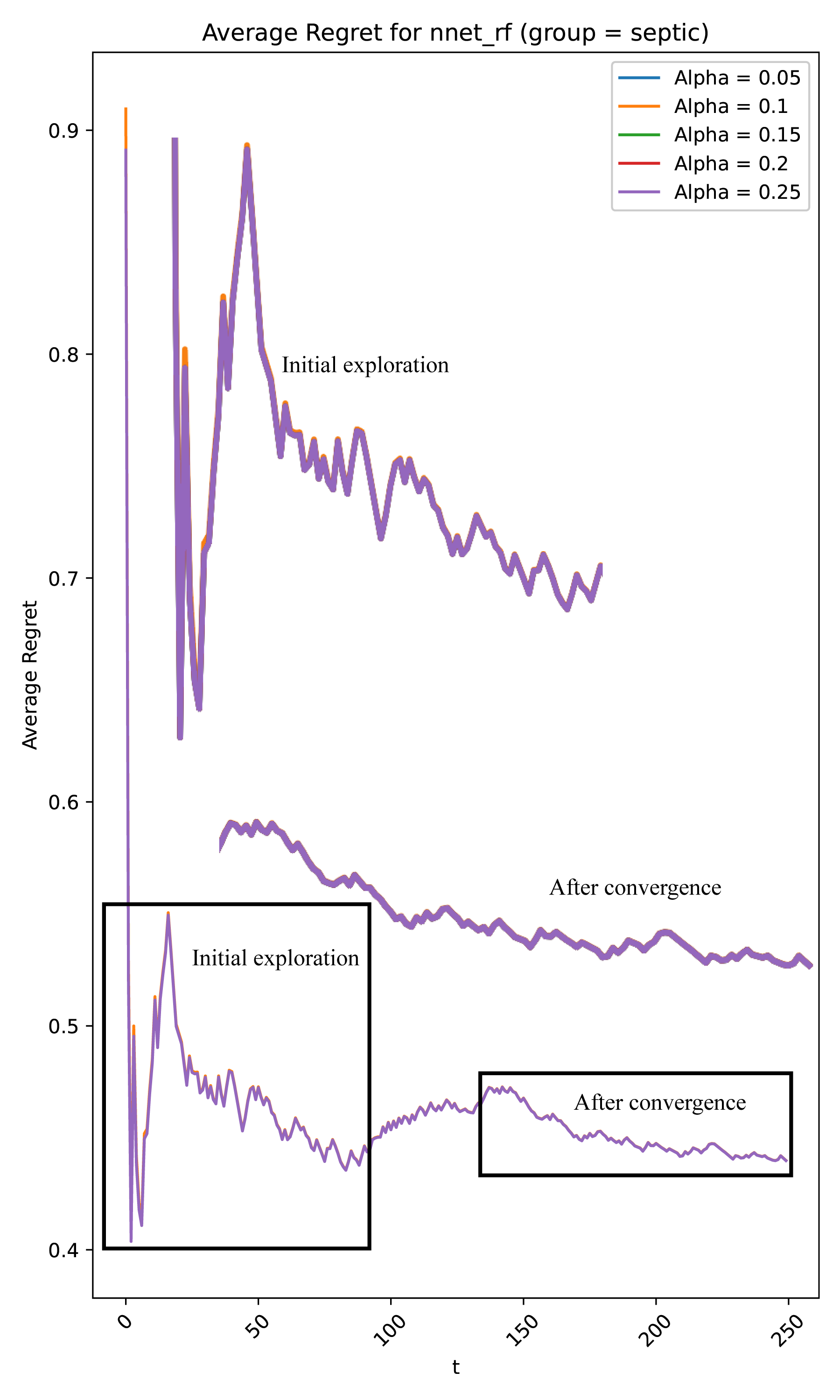}
    \caption{Average Regret (nnet\_rf).}
    \label{fig:AR_nnet_rf_differnt_alphas}
     \end{subfigure}
     \begin{subfigure}{0.4\textwidth}
         \centering
    \includegraphics[width=\linewidth]{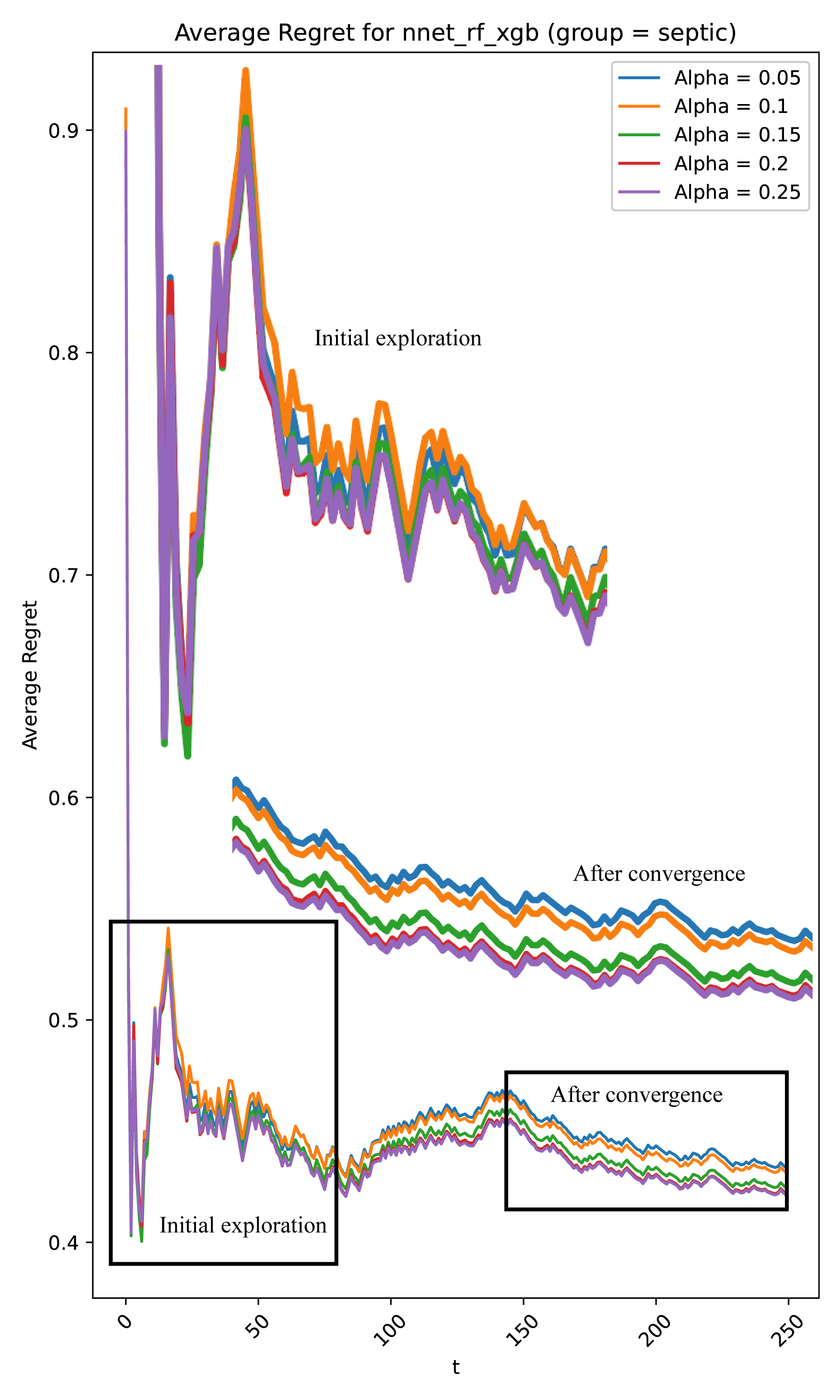}
    \caption{Average Regret (nnet\_rf\_xgb).}
    \label{fig:AR_nnet_rf_xgb_differnt_alphas}
     \end{subfigure}
     \caption{Average Regret of Different $\alpha$ (baseline = nnet).}
\end{figure*}
The rf model provides stability and robustness, while the xgb model captures complex patterns. When combined, these models cover a wide range of predictive capabilities, ensuring that the overall ensemble can handle both simpler and more complex cases effectively. This can explain why the average regret of combination rf\_xgb is the lowest among all combinations when 
$\alpha = 0.05$ (see \ref{fig:AR_rfcombo005}) and $\alpha = 0.1$ (see \ref{fig:AR_rfcombo01}). 

 \begin{figure*}[h!]
     \centering
     \begin{subfigure}{0.4\textwidth}
         \centering
    \includegraphics[width=\linewidth]{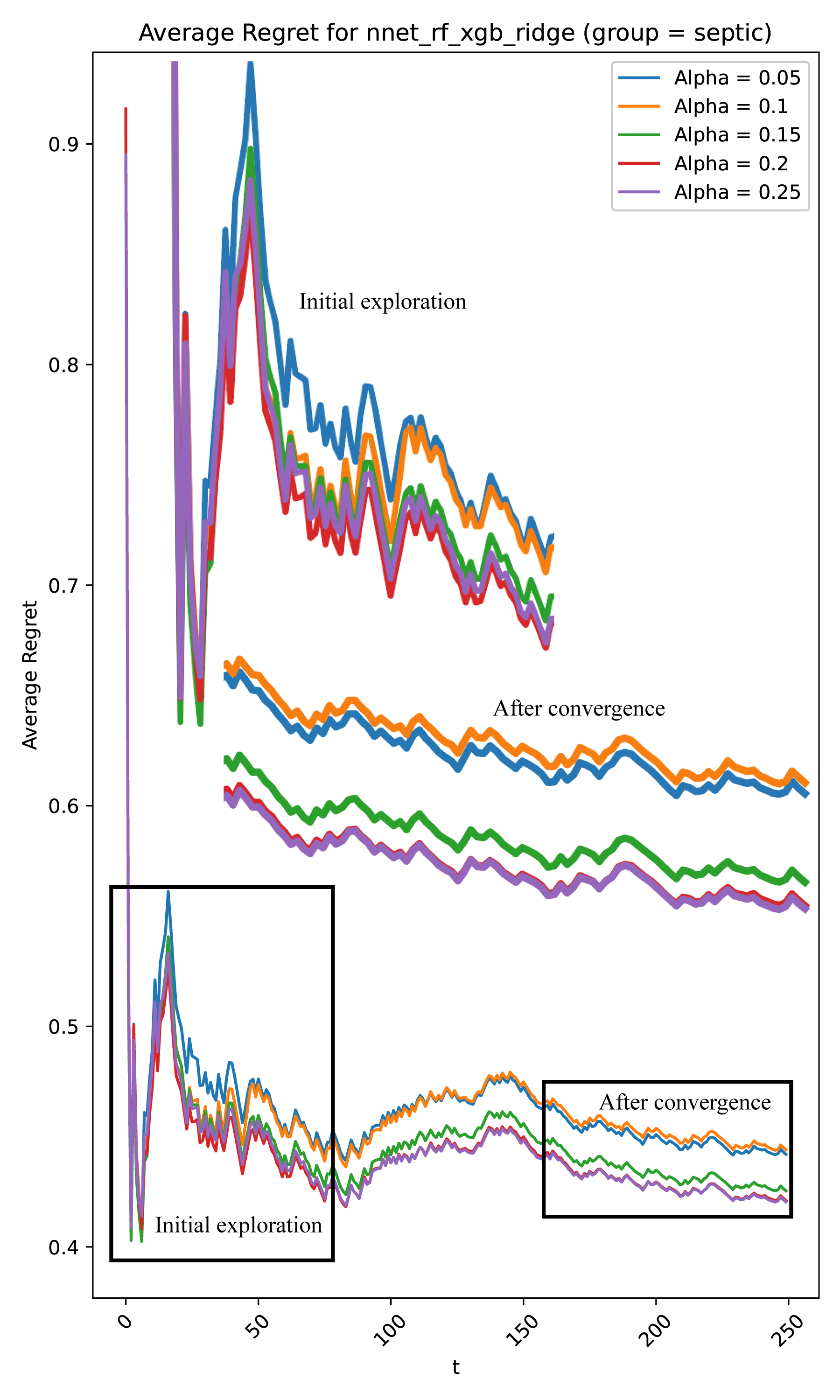}
   \caption{Average Regret (nnet\_rf\_xgb\_ridge).}
    \label{fig:AR_nnet_rf_xgb_ridge_differnt_alphas}
     \end{subfigure}
     \begin{subfigure}{0.4\textwidth}
         \centering
    \includegraphics[width=\linewidth]{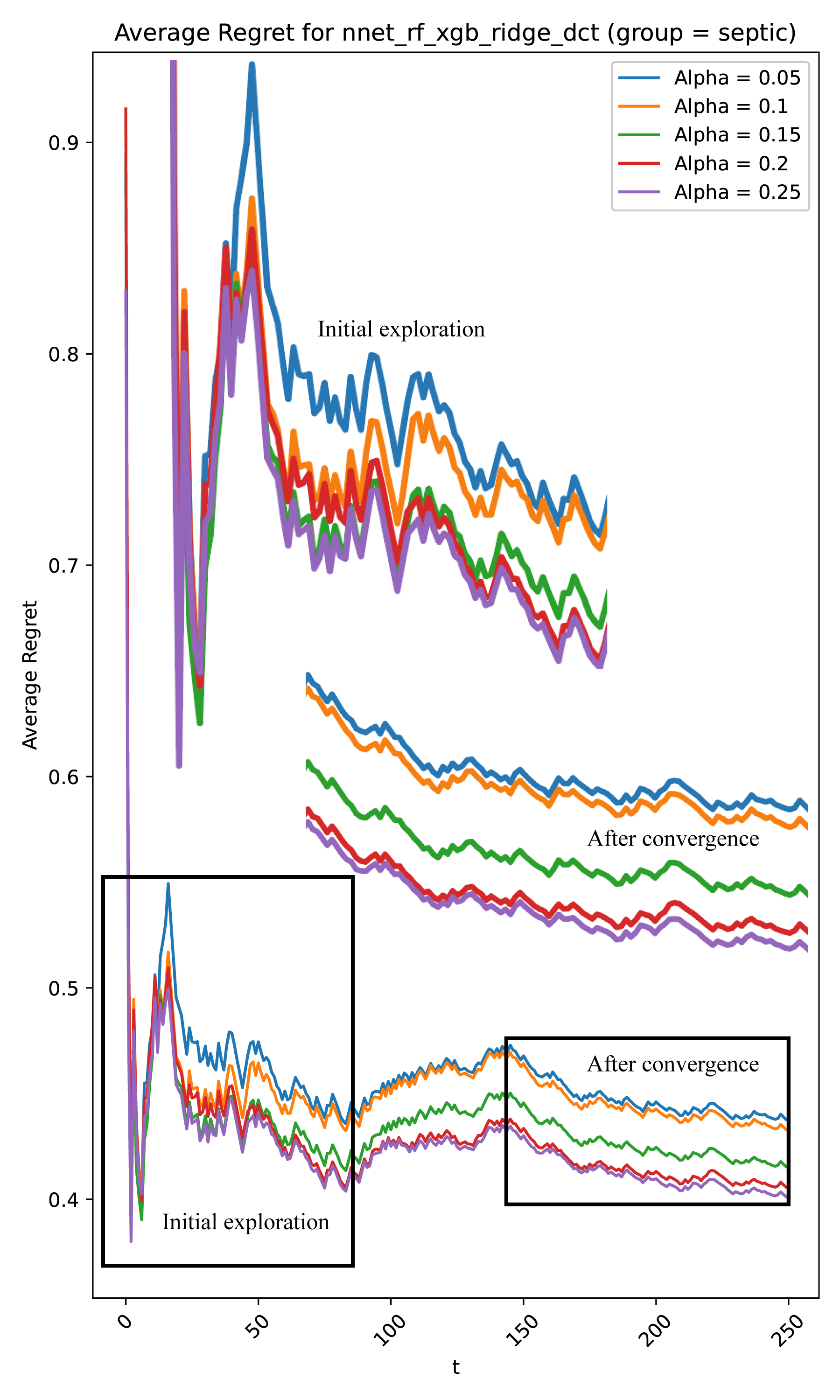}
        \caption{Average Regret (nnet\_rf\_xgb\_ridge\_dct).}
    \label{fig:AR_nnet_rf_xgb_ridge_dct_differnt_alphas}
     \end{subfigure}
     \caption{Average Regret of Different $\alpha$ (baseline = nnet).}
\end{figure*}

With lower $\alpha$ values, the Sepsyn-OLCP algorithm favors exploitation over exploration. Since the rf\_xgb combination provides a well-balanced approach to learning (stability from rf and complexity from xgb as aforementioned), it efficiently uses the historical data to make accurate predictions, minimizing regret.

The rf\_xgb combination is likely already well-calibrated, meaning that the predictions made by this combination are sufficiently accurate to require less exploration. Thus, narrower confidence intervals (lower $\alpha$) help the model make confident and effective decisions, reducing the average regret.

The ensemble benefits from reduced variance (due to rf) and reduced bias (due to xgb). This balance ensures that the model can generalize well across different patient cases, leading to lower regret, especially when exploration is limited.
The rf\_xgb combination may be particularly well-suited to the dataset's characteristics, allowing it to outperform other combinations under conservative exploration settings.
Furthermore, since the dataset has both linear and nonlinear components, the rf\_xgb combination can better capture these relationships, making it more effective even when the exploration budget is low (as determined by lower $\alpha$).

In conclusion, the proposed framework's performance is determined by the significance level and the performance of each individual AI candidate.

\begin{table}[htb!]
\centering
\scalebox{1}{\begin{tabular}{lcccccc}
\hline
\hline
Experts & $\alpha$ & AUROC & AUPRC & Accuracy & F-measure & Utility \\
\hline
nnet & 0.05 & 0.6488 & 0.1759 & 0.8884 & 0.0131 & 0.0061 \\
nnet & 0.1 & 0.6488 & 0.1759 & 0.8884 & 0.0131 & 0.0061 \\
nnet & 0.15 & 0.6488 & 0.1759 & 0.8884 & 0.0131 & 0.0061 \\
nnet & 0.2 & 0.6488 & 0.1759 & 0.8884 & 0.0131 & 0.0061 \\
nnet & 0.25 & 0.6488 & 0.1759 & 0.8884 & 0.0131 & 0.0061 \\
\hline
nnet, rf & 0.05 & 0.6489 & 0.1751 & 0.8881 & 0.0163 & 0.0072 \\
nnet, rf & 0.1 & 0.6481 & 0.1747 & 0.8881 & 0.0163 & 0.0072 \\
nnet, rf & 0.15 & 0.6489 & 0.1751 & 0.8881 & 0.0163 & 0.0072 \\
nnet, rf & 0.2 & 0.6489 & 0.1751 & 0.8881 & 0.0163 & 0.0072 \\
nnet, rf & 0.25 & 0.6489 & 0.1751 & 0.8881 & 0.0163 & 0.0072 \\
\hline
nnet, rf, xgb & 0.05 & 0.6723 & 0.2015 & 0.8901 & 0.0034 & 0.0009 \\
nnet, rf, xgb & 0.1 & 0.6837 & 0.2170 & 0.8902 & 0.0067 & 0.0025 \\
nnet, rf, xgb & 0.15 & 0.7262 & 0.2734 & 0.8906 & 0.0067 & 0.0021 \\
nnet, rf, xgb & 0.2 & 0.7182 & 0.2577 & 0.8911 & 0.0151 & 0.0046 \\
nnet, rf, xgb & 0.25 & 0.7205 & 0.2737 & 0.8911 & 0.0183 & 0.0055 \\
\hline
nnet, rf, xgb, ridge & 0.05 & 0.6586 & 0.1851 & 0.8903 & 0.0008 & 0.0003 \\
nnet, rf, xgb, ridge & 0.1 & 0.6626 & 0.1869 & 0.8901 & 0.0050 & 0.0019 \\
nnet, rf, xgb, ridge & 0.15 & 0.7232 & 0.2982 & 0.8905 & 0.0042 & 0.0012 \\
nnet, rf, xgb, ridge & 0.2 & 0.7215 & 0.2980 & 0.8912 & 0.0167 & 0.0050 \\
nnet, rf, xgb, ridge & 0.25 & 0.7246 & 0.3099 & 0.8911 & 0.0151 & 0.0046 \\
\hline
nnet, rf, xgb, ridge, dct & 0.05 & 0.6917 & 0.2090 & 0.8901 & 0.0541 & 0.0255 \\
nnet, rf, xgb, ridge, dct & 0.1 & 0.6875 & 0.2105 & 0.8899 & 0.0839 & 0.0404 \\
nnet, rf, xgb, ridge, dct & 0.15 & 0.7331 & 0.2668 & 0.8924 & 0.0681 & 0.0307 \\
nnet, rf, xgb, ridge, dct & 0.2 & 0.7357 & 0.3255 & 0.8954 & 0.1002 & 0.0446 \\
nnet, rf, xgb, ridge, dct & 0.25 & 0.7340 & 0.3573 & 0.8952 & 0.0885 & 0.0370 \\
\hline
\hline
\end{tabular}}
\caption{Performance Metrics for nnet Baseline.}
\label{tab:results_nnet_1000_250}
\end{table}

\begin{table}[htb!]
\centering
 \scalebox{1}{\begin{tabular}{lcccccc}
\hline
\hline
Experts & $\alpha$ & AUROC & AUPRC & Accuracy & F-measure & Utility \\
\hline
xgb & 0.05 & 0.7167 & 0.2645 & 0.8915 & 0.0225 & 0.0069 \\
xgb & 0.1 & 0.7167 & 0.2645 & 0.8915 & 0.0225 & 0.0069 \\
xgb & 0.15 & 0.7167 & 0.2645 & 0.8915 & 0.0225 & 0.0069 \\
xgb & 0.2 & 0.7167 & 0.2645 & 0.8915 & 0.0225 & 0.0069 \\
xgb & 0.25 & 0.7167 & 0.2645 & 0.8915 & 0.0225 & 0.0069 \\
\hline
xgb, rf & 0.05 & 0.7162 & 0.2644 & 0.8915 & 0.0225 & 0.0069 \\
xgb, rf & 0.1 & 0.7162 & 0.2619 & 0.8915 & 0.0217 & 0.0066 \\
xgb, rf & 0.15 & 0.7161 & 0.2618 & 0.8915 & 0.0217 & 0.0066 \\
xgb, rf & 0.2 & 0.7169 & 0.2652 & 0.8915 & 0.0225 & 0.0069 \\
xgb, rf & 0.25 & 0.7170 & 0.2654 & 0.8915 & 0.0225 & 0.0069 \\
\hline
xgb, rf, lr & 0.05 & 0.6609 & 0.2002 & 0.8904 & 0.0017 & 0.0005 \\
xgb, rf, lr & 0.1 & 0.6756 & 0.2036 & 0.8904 & 0.0008 & 0.0003 \\
xgb, rf, lr & 0.15 & 0.6996 & 0.2871 & 0.8911 & 0.0142 & 0.0044 \\
xgb, rf, lr & 0.2 & 0.7198 & 0.3337 & 0.8916 & 0.0225 & 0.0069 \\
xgb, rf, lr & 0.25 & 0.7079 & 0.3398 & 0.8916 & 0.0225 & 0.0069 \\
\hline
xgb, rf, dct, lr & 0.05 & 0.6895 & 0.2242 & 0.8905 & 0.0573 & 0.0264 \\
xgb, rf, dct, lr & 0.1 & 0.6778 & 0.1940 & 0.8864 & 0.0795 & 0.0401 \\
xgb, rf, dct, lr & 0.15 & 0.7223 & 0.2565 & 0.8905 & 0.0982 & 0.0455 \\
xgb, rf, dct, lr & 0.2 & 0.7328 & 0.3320 & 0.8939 & 0.0969 & 0.0428 \\
xgb, rf, dct, lr & 0.25 & 0.7414 & 0.3920 & 0.8948 & 0.0896 & 0.0373 \\
\hline
xgb, cat, rf, dct, lr & 0.05 & 0.6891 & 0.2091 & 0.8899 & 0.0388 & 0.0169 \\
xgb, cat, rf, dct, lr & 0.1 & 0.6834 & 0.1954 & 0.8876 & 0.0802 & 0.0386 \\
xgb, cat, rf, dct, lr & 0.15 & 0.7348 & 0.2789 & 0.8922 & 0.0797 & 0.0352 \\
xgb, cat, rf, dct, lr & 0.2 & 0.7398 & 0.3398 & 0.8940 & 0.0728 & 0.0303 \\
xgb, cat, rf, dct, lr & 0.25 & 0.7331 & 0.3793 & 0.8942 & 0.0699 & 0.0290 \\
\hline
\hline
\end{tabular}}
\caption{Performance Metrics for xgb Baseline.}
\label{tab:results_xgb_1000_250}
\end{table}

 \begin{table}[htb!]
\centering
 \scalebox{1}{\begin{tabular}{lcccccc}
\hline
\hline
Experts & $\alpha$ & AUROC & AUPRC & Accuracy & F-measure & Utility \\
\hline
rf & 0.05 & 0.6743 & 0.2079 & 0.8903 & 0.0000 & 0.0000 \\
rf & 0.1 & 0.6743 & 0.2079 & 0.8903 & 0.0000 & 0.0000 \\
rf & 0.15 & 0.6743 & 0.2079 & 0.8903 & 0.0000 & 0.0000 \\
rf & 0.2 & 0.6743 & 0.2079 & 0.8903 & 0.0000 & 0.0000 \\
rf & 0.25 & 0.6743 & 0.2079 & 0.8903 & 0.0000 & 0.0000 \\
\hline
rf, xgb & 0.05 & 0.7154 & 0.2629 & 0.8915 & 0.0225 & 0.0069 \\
rf, xgb & 0.1 & 0.7154 & 0.2604 & 0.8915 & 0.0217 & 0.0066 \\
rf, xgb & 0.15 & 0.7156 & 0.2611 & 0.8915 & 0.0217 & 0.0066 \\
rf, xgb & 0.2 & 0.7163 & 0.2639 & 0.8915 & 0.0225 & 0.0069 \\
rf, xgb & 0.25 & 0.7163 & 0.2639 & 0.8915 & 0.0225 & 0.0069 \\
\hline
rf, xgb, lr & 0.05 & 0.6599 & 0.2014 & 0.8904 & 0.0017 & 0.0005 \\
rf, xgb, lr & 0.1 & 0.6867 & 0.1987 & 0.8904 & 0.0008 & 0.0003 \\
rf, xgb, lr & 0.15 & 0.7227 & 0.2935 & 0.8911 & 0.0151 & 0.0046 \\
rf, xgb, lr & 0.2 & 0.7184 & 0.3279 & 0.8916 & 0.0225 & 0.0069 \\
rf, xgb, lr & 0.25 & 0.7021 & 0.3312 & 0.8916 & 0.0225 & 0.0069 \\
\hline
rf, xgb, ridge, lr & 0.05 & 0.6576 & 0.1946 & 0.8903 & 0.0000 & 0.0000 \\
rf, xgb, ridge, lr & 0.1 & 0.6857 & 0.1997 & 0.8903 & 0.0000 & 0.0000 \\
rf, xgb, ridge, lr & 0.15 & 0.6876 & 0.2545 & 0.8905 & 0.0050 & 0.0015 \\
rf, xgb, ridge, lr & 0.2 & 0.7086 & 0.3168 & 0.8915 & 0.0225 & 0.0069 \\
rf, xgb, ridge, lr & 0.25 & 0.7068 & 0.3114 & 0.8915 & 0.0217 & 0.0066 \\
\hline
rf, xgb, cat, dct, lr & 0.05 & 0.6891 & 0.2091 & 0.8899 & 0.0388 & 0.0169 \\
rf, xgb, cat, dct, lr & 0.1 & 0.6834 & 0.1954 & 0.8876 & 0.0802 & 0.0386 \\
rf, xgb, cat, dct, lr & 0.15 & 0.7348 & 0.2789 & 0.8922 & 0.0797 & 0.0352 \\
rf, xgb, cat, dct, lr & 0.2 & 0.7398 & 0.3398 & 0.8940 & 0.0728 & 0.0303 \\
rf, xgb, cat, dct, lr & 0.25 & 0.7331 & 0.3793 & 0.8942 & 0.0699 & 0.0290 \\
\hline
\hline
\end{tabular}}
\caption{Performance Metrics for rf Baseline.}
\label{tab:results_rf_1000_250}
\end{table}
\section{Discussion}
\label{sec:disc}
The introduction of Sepsyn-OLCP, an online learning-based framework leveraging conformal prediction guarantees, demonstrates significant potential for enhancing early sepsis prediction in clinical settings. The framework, with its robust theoretical underpinnings, a conformal prediction mechanism for uncertainty quantification, and a gap-based Bayesian bandit model, is designed to adapt decision-making dynamically when new data is available. 
In this section, we highlight the key findings, insights, and implications from the experimental results, underscoring the robustness of Sepsyn-OLCP.


Sepsyn-OLCP proved effective in leveraging a pool of AI clinicians to achieve better predictive performance over time compared to standalone models. By integrating multiple models such as neural networks (nnet), random forests (rf), and gradient boosting models (xgb), the framework dynamically balanced exploration (i.e., testing underutilized models) and exploitation (i.e., utilizing high-performing models). The observed decrease in average regret across combinations (e.g., rf\_xgb, nnet\_rf) highlights the framework’s ability to minimize decision-making errors while maintaining robustness. Notably, the addition of complementary models (e.g., random forests with XGBoost) yielded lower regrets due to their synergistic strengths—random forests provided stability while XGBoost captured complex patterns.

The significance level $\alpha$, which determines the width of confidence intervals in conformal prediction, played a critical role in controlling the balance between exploration and exploitation. Lower $\alpha$ values (e.g., 0.05, 0.1) resulted in narrower intervals and more conservative decision-making, leading to stable yet potentially less adaptive outcomes. Conversely, higher $\alpha$ values (e.g., 0.2, 0.25) allowed for broader intervals and more aggressive exploration, enabling the algorithm to uncover potentially high-performing models at the cost of increased initial regret.

From the experimental results, combinations like rf\_xgb consistently showed improved performance at lower $\alpha$, while larger ensembles (e.g., rf\_xgb\_ridge\_lr\_dct) demonstrated greater performance variability across $\alpha$. This indicates that smaller, well-calibrated combinations benefit from conservative exploration, while larger combinations require more aggressive exploration to identify optimal subsets of candidate models.


The experimental results highlight the power of combining models with complementary characteristics. For example,
\begin{itemize}
    \item \textbf{Random Forest and XGBoost (rf\_xgb):} This combination consistently minimized average regret due to its balanced nature. Random forests offered stability by aggregating predictions from decision trees, while XGBoost excelled at capturing intricate, nonlinear patterns.
    \item \textbf{Inclusion of Logistic Regression and Ridge Regression:} Adding linear models like logistic regression (lr) and ridge regression to ensembles (e.g., rf\_xgb\_lr, rf\_xgb\_ridge\_lr) diversified the model pool, enhancing the framework's ability to generalize across linear and nonlinear data structures. This was particularly effective at moderate $\alpha$ levels.
   \item \textbf{Complex Ensembles (e.g., xgb\_cat\_rf\_dct\_lr):} Large ensembles exhibited increased potential for high predictive performance but required higher $\alpha$ values to enable adequate exploration. Despite higher initial regrets, these combinations achieved notable performance improvements in terms of AUROC and utility scores.
\end{itemize}

\subsection{AUROC, AUPRC, F-measure, and Utility Score Performance}
Key performance metrics such as AUROC, AUPRC, F-measure, and utility scores varied significantly depending on the model combinations and $\alpha$ values.

AUROC values help in evaluating how well the model discriminates between positive (sepsis onset) and negative cases (no sepsis).
Across different model combinations, we see AUROC values ranging from 0.6488 (nnet alone) to 0.7357 (nnet, rf, xgb, ridge, dct at $\alpha= 0.2$).
In the context of early sepsis prediction, AUROC values closer to 0.7 indicate that the ensemble with more diverse models (such as the full combination of five models) performs better in distinguishing between patients at risk of sepsis and those not at risk. However, given the critical nature of sepsis prediction, an AUROC around 0.7 still indicates room for improvement.


AUPRC is highly valuable in the context of imbalanced data, as it focuses on the balance between precision (minimizing false positives) and recall (capturing as many true positives as possible).
The AUPRC steadily increases as more models are combined, reaching 0.3573 for the full ensemble (nnet, rf, xgb, ridge, dct) at $\alpha= 0.25$. This is a significant improvement compared to using nnet alone (0.1759).
The improvement in AUPRC when adding models such as xgb, ridge, and dct suggests that these models effectively capture more true positives while managing false positives, which is crucial in sepsis prediction to avoid missing potential sepsis cases.

The accuracy metric stays fairly high across all models, ranging from 0.8884 (nnet alone) to 0.8954 (nnet, rf, xgb, ridge, dct at $\alpha = 0.2$).
In sepsis prediction, achieving high accuracy may mean correctly predicting non-sepsis cases, but it does not guarantee that the critical sepsis cases are detected. while accuracy is high, it is not necessarily the most informative metric in this context due to class imbalance; it is likely that most predictions are for non-sepsis events, which inflates the accuracy value. However, in our experiments, the ratio of septic and non-septic patients is 1:1, so even with a 1:1 ratio, we can still achieve high accuracy for the whole cohort. 

The F-measure (harmonic mean of precision and recall) is particularly important for understanding how well the model handles the balance between precision (avoiding false positives) and recall (capturing true positives).
For nnet alone, the F-measure is very low (0.0131), indicating poor performance in detecting sepsis events despite high accuracy. This is indicative of the challenges of class imbalance and false positives that reduce the precision of the model.
Adding dct (decision tree) results in substantial improvements in F-measure, reaching 0.1002 for the full ensemble at $\alpha = 0.2$. This indicates a better balance between correctly predicting true positives and minimizing false alarms, which is particularly critical in clinical applications where unnecessary interventions need to be minimized.

Individual models like nnet (see \ref{tab:results_nnet_1000_250}), rf (see \ref{tab:results_rf_1000_250}), and xgb (see \ref{tab:results_xgb_1000_250}) demonstrated stable but limited performance, with AUROCs ranging from 0.65 to 0.72 and utility scores near 0. Standalone decision trees (dct) performed well initially. However, the performance of individual models cannot compete with that of multiple AI clinicians due to the limited adaptability of individual models.

Combinations like rf\_xgb and rf\_xgb\_lr consistently achieved higher AUROCs and utility scores, which is consistent with the regret analysis. Larger ensembles such as xgb\_cat\_rf\_dct\_lr achieved the best utility scores and F-measures, particularly at $\alpha = 0.2$ and $\alpha = 0.25$, indicating the importance of exploration in complex setups.

Sepsyn-OLCP's adaptability to diverse clinical scenarios is a key strength. Its ability to process incoming patient data in real time makes it a perfect fit for clinical environments where data distributions can shift and patient conditions are diverse. The use of conformal prediction ensures rigorous uncertainty quantification, empowering clinicians to interpret predictions with confidence. For instance, prediction intervals allow clinicians to assess not just the most probable outcomes but also the associated uncertainties, which is crucial for high-stakes decisions like sepsis management.
\begin{table}[htb!]
\centering
 \scalebox{1}{\begin{tabular}{lcccccc}
\hline
\hline
Experts & $\alpha$ & AUROC & AUPRC & Accuracy & F-measure & Utility \\
\hline
dct & 0.05 & 0.6950 & 0.2520 & 0.8875 & 0.1578 & 0.0800 \\
dct & 0.1 & 0.6950 & 0.2520 & 0.8875 & 0.1578 & 0.0800 \\
dct & 0.15 & 0.6950 & 0.2520 & 0.8875 & 0.1578 & 0.0800 \\
dct & 0.2 & 0.6950 & 0.2520 & 0.8875 & 0.1578 & 0.0800 \\
dct & 0.25 & 0.6950 & 0.2520 & 0.8875 & 0.1578 & 0.0800 \\
\hline
dct, cat & 0.05 & 0.6950 & 0.2520 & 0.8875 & 0.1578 & 0.0800 \\
dct, cat & 0.1 & 0.6951 & 0.2514 & 0.8875 & 0.1572 & 0.0798 \\
dct, cat & 0.15 & 0.6950 & 0.2520 & 0.8875 & 0.1578 & 0.0800 \\
dct, cat & 0.2 & 0.6952 & 0.2523 & 0.8875 & 0.1578 & 0.0800 \\
dct, cat & 0.25 & 0.6951 & 0.2521 & 0.8875 & 0.1578 & 0.0800 \\
\hline
dct, rf, cat & 0.05 & 0.7065 & 0.2347 & 0.8888 & 0.1016 & 0.0474 \\
dct, rf, cat & 0.1 & 0.6983 & 0.2187 & 0.8879 & 0.1055 & 0.0507 \\
dct, rf, cat & 0.15 & 0.7119 & 0.2602 & 0.8919 & 0.1290 & 0.0621 \\
dct, rf, cat & 0.2 & 0.7330 & 0.3143 & 0.8947 & 0.1273 & 0.0592 \\
dct, rf, cat & 0.25 & 0.7417 & 0.3425 & 0.8956 & 0.1200 & 0.0540 \\
\hline
dct, rf, ridge, cat & 0.05 & 0.6938 & 0.2219 & 0.8905 & 0.0714 & 0.0331 \\
dct, rf, ridge, cat & 0.1 & 0.6762 & 0.1932 & 0.8874 & 0.0890 & 0.0438 \\
dct, rf, ridge, cat & 0.15 & 0.7140 & 0.2475 & 0.8915 & 0.1059 & 0.0492 \\
dct, rf, ridge, cat & 0.2 & 0.7404 & 0.3268 & 0.8948 & 0.1047 & 0.0469 \\
dct, rf, ridge, cat & 0.25 & 0.7386 & 0.3642 & 0.8956 & 0.1012 & 0.0437 \\
\hline
dct, ridge, cat, rf, lr & 0.05 & 0.6790 & 0.2036 & 0.8904 & 0.0497 & 0.0234 \\
dct, ridge, cat, rf, lr & 0.1 & 0.6815 & 0.1926 & 0.8874 & 0.0738 & 0.0357 \\
dct, ridge, cat, rf, lr & 0.15 & 0.7208 & 0.2503 & 0.8918 & 0.0758 & 0.0345 \\
dct, ridge, cat, rf, lr & 0.2 & 0.7337 & 0.3342 & 0.8938 & 0.0734 & 0.0303 \\
dct, ridge, cat, rf, lr & 0.25 & 0.7269 & 0.3580 & 0.8943 & 0.0737 & 0.0305 \\
\hline
\hline
\end{tabular}}
\caption{Performance Metrics for dct Baseline.}
\label{tab:results_dct_1000_250}
\end{table}
A critical insight from the experiments is the trade-off between conservative exploration (low $\alpha$) and aggressive exploration (high $\alpha$). Conservative exploration minimizes initial regret but risks missing high-performing models. Aggressive exploration incurs higher initial regret but can identify superior models over time. This trade-off underscores the need for adaptive $\alpha$ tuning based on the clinical context and available data, providing a practical insight for future research and real-world deployment.

\section{Conclusion}
\label{sec:conc}
 
While Sepsyn-OLCP demonstrates robust performance, there are areas for improvement:
\begin{itemize}
    \item \textbf{Scalability to Larger Ensembles:} Larger ensembles introduced higher computational complexity. Future work could explore pruning mechanisms to identify optimal model subsets dynamically.
    \item \textbf{Dynamic $\alpha$ Adjustment:} 
    The static choice of $\alpha$ limits adaptability. Incorporating dynamic $\alpha$ adjustment based on real-time performance feedback could enhance exploration-exploitation balance.
    \item \textbf{Incorporation of Temporal Dynamics: }
    The current framework primarily focuses on static contexts. Extending the model to capture temporal patterns in patient data could improve predictions, particularly for time-sensitive conditions like sepsis.
    \item \textbf{Broader Evaluation Metrics:} While AUROC and AUPRC are standard metrics, incorporating metrics that reflect clinical impact, such as treatment timeliness and outcome improvement, would provide more actionable insights.
\end{itemize}

Sepsyn-OLCP exemplifies how online learning and conformal prediction can be combined to address complex clinical decision-making tasks like early sepsis prediction. By leveraging multiple AI clinicians and dynamically balancing exploration and exploitation, the framework achieves both predictive accuracy and robustness. The experimental results underscore its versatility across diverse model combinations and clinical scenarios, laying the groundwork for further research and real-world deployment.

\bibliographystyle{unsrt}  
\bibliography{references}

\end{document}